\documentclass[conference]{IEEEtran}
\IEEEoverridecommandlockouts
\usepackage{cite}
\usepackage{amsmath,amssymb,amsfonts}
\usepackage{algorithmic}
\usepackage{graphicx}
\usepackage{textcomp}
\usepackage{cleveref}
\usepackage{xcolor}
\usepackage{adjustbox}

\usepackage{algorithmic}
\usepackage{booktabs}
\usepackage{siunitx}
\usepackage{caption}
\usepackage{subcaption}

\newcommand{\ba}{\mathbf{a}}
\newcommand{\bb}{\mathbf{b}}
\newcommand{\be}{\mathbf{e}}
\newcommand{\bh}{\mathbf{h}}
\newcommand{\bm}{\mathbf{m}}
\newcommand{\bp}{\mathbf{p}}

\newcommand{\bE}{\mathbf{E}}
\newcommand{\bH}{\mathbf{H}}
\newcommand{\bP}{\mathbf{P}}
\newcommand{\bW}{\mathbf{W}}

\newcommand{\rb}[1]{\left(#1\right)}
\def\BibTeX{{\rm B\kern-.05em{\sc i\kern-.025em b}\kern-.08em
    T\kern-.1667em\lower.7ex\hbox{E}\kern-.125emX}}

\begin{document}

\title{Graph Neural Network Architecture Search for \\ Molecular Property Prediction
}

\author{\IEEEauthorblockN{Shengli Jiang}
\IEEEauthorblockA{\textit{Department of Chemical and Biological Engineering} \\
\textit{University of Wisconsin {\textemdash} Madison}\\
Madison, USA \\
sjiang87@wisc.edu}
\and
\IEEEauthorblockN{Prasanna Balaprakash}
\IEEEauthorblockA{\textit{Mathematics and Computer Science Division \&} \\
\textit{Leadership Computing Facility} \\
\textit{Argonne National Laboratory}\\
Lemont, USA \\
pbalapra@anl.gov}
}

\maketitle

\begin{abstract}

Predicting the properties of a molecule from its structure is a challenging task.  
Recently, deep learning methods have improved the state of the art for this task because of their ability to learn useful features from the given data. By treating molecule structure as graphs, where atoms and bonds are modeled as nodes and edges, graph neural networks (GNNs) have been widely used to predict molecular properties. However, the design and development of GNNs for a given dataset rely on labor-intensive design and tuning of the network architectures. Neural architecture search (NAS) is a promising approach to discover high-performing neural network architectures automatically. To that end, we develop an NAS approach to automate the design and development of GNNs for molecular property prediction. 
Specifically, we focus on automated development of message-passing neural networks (MPNNs) to predict the molecular properties of small molecules in quantum mechanics and physical chemistry datasets from the MoleculeNet benchmark. 
We demonstrate the superiority of the automatically discovered MPNNs by comparing them with manually designed GNNs from the MoleculeNet benchmark. We study the relative importance of the choices in the MPNN search space, demonstrating that customizing the architecture is critical to enhancing performance in molecular property prediction and that the proposed approach can perform customization automatically with minimal manual effort.

\end{abstract}

\begin{IEEEkeywords}
graph neural networks, neural architecture search, regularized evolution, deep learning, AutoML
\end{IEEEkeywords}

\section{Introduction}


Molecular property prediction is an important task in science and engineering. In pharmaceutical science, predicting the drug molecule properties can enable the design of next-generation drugs \cite{burbidge2001drug}. In material science, predicting the band gap of perovskites is essential in the design of novel solar cells \cite{tao2017accurate}. In electrochemistry, 
finding new electrolytes with high Li-ion conductivity will speed up solid-state Li-ion battery research \cite{sendek2018machine}. Traditionally, simulation-based methods, such as quantum Monte Carlo (QMC) \cite{mascagni2004monte}, molecular dynamics (MD) \cite{varshney2008molecular}, and density functional theory (DFT) \cite{becke2014perspective}, are used to predict molecular properties. QMC and MD consider atoms as classical particles and simulate the interactions between atoms or molecules by using empirical potential forms such as Lennard-Jones. DFT can predict the quantum properties of molecules by calculating the ground-state energy of systems using exchange correlation functions \cite{becke2014perspective}. Although they achieve high prediction accuracy, the traditional methods are computationally expensive \cite{lu2019molecular}. Prior studies have shown that predicting the property of a molecule with 20 atoms using DFT can take an hour \cite{gilmer2017neural}. 

In the past decade, because of the availability of cheap computational power and increased coordinated data-collection efforts, a wide range of molecular datasets have been generated by simulations and experiments \cite{blum}. However, developing accurate predictive models for these molecular datasets remains a difficult task. For a molecular propriety dataset, the amount of training data is often limited by the expense of simulation and/or experiments. The molecular datasets are  diverse in terms of their molecular structure and properties. Consequently, a model trained for one molecular dataset cannot be transferred to another because of the non-Euclidean characteristics of the molecular structure data \cite{wu2018moleculenet}.



Graph neural networks (GNNs) have been widely used to predict molecular properties \cite{gilmer2017neural, wu2018moleculenet}. By treating molecules as graphs, where nodes are atoms and edges are bonds, GNNs capture the non-Euclidean nature of molecules. Message-passing neural networks (MPNNs) \cite{gilmer2017neural}, a class of GNNs, provide a generic framework to incorporate complex node (atomic) and edge (bond) features. The common node features include the mass number, the atomic number, and the number of valence electrons. The common edge features comprise the bond type (e.g., single or double bond), the bond length, and whether a bond is in a ring. Following the concepts of MPNNs, node features can be passed as \textit{messages} from one node to another along edges. Researchers can discern the hidden feature of a node by iteratively aggregating features from neighboring nodes and the node itself. Once identified, the formerly hidden features can be used by other machine learning techniques (e.g., fully connected neural networks) to perform specific tasks (e.g., classification, regression, and clustering). The advantage of MPNNs compared with other GNNs can be attributed to their ability to integrate edge features into the message passing of node features. Specifically, edge features are processed by a neural network to generate some weights to guide the message passing between nodes.

Despite the superior performance of MPNNs, arduous tuning of the network architecture impedes their development and wider adoption. To obtain high prediction accuracy for different datasets, we need to tune various MPNN components, including message, aggregate, update, and gather (readout) functions. For example, GG-NN \cite{li2015gated} uses a gated graph unit as the update function, whereas an Interaction Network \cite{battaglia2016interaction} simply updates with a dense neural network. The manually designed MPNNs not only require a substantial number of experiments in the design space but also tend to be lower-performing when applied to a new dataset. The data-specific nature of MPNN design means that it urgently needs an automated MPNN search to identify the best task-specific architecture for a given dataset.

Neural architecture search (NAS) has been designed to automatically search for the best network architecture for a given dataset. Specifically, it uses a search method  (e.g., reinforcement learning, evolutionary algorithm, or stochastic gradient descent) to explore the user-defined search space and chooses the best architecture based on the performance of the generated model (e.g., validation accuracy) on a given task. The search space contains all possible architectures. The architectures discovered by NAS have been shown to outperform manually designed architectures in various tasks such as image classification \cite{real2017large}, image segmentation \cite{liu2019auto}, natural language processing \cite{fan2020searching}, and time series prediction \cite{maulik2020recurrent}. Despite their impact, NAS methods for GNNs and, in particular, MPNNs have received relatively little attention in the literature. 
In two recent studies \cite{zhou2019auto, gao2019graphnas}, NAS methods were presented that  generate GNNs for classifying node properties. In these studies, however, the NAS-GNN search space lacks the integration of edge features, which are crucial for molecular property prediction \cite{gilmer2017neural}.

We focus on developing NAS for MPNNs that incorporates both the node and edge features to predict molecular properties. MPNNs have been widely used to study molecular properties \cite{gilmer2017neural}. Borrowing the idea of Res-Net \cite{szegedy2017inception}, we develop a NAS search space for stacked MPNNs with multiple MPNN cells and skip connections. We implement our approach in DeepHyper \cite{balaprakash2018deephyper} {\textemdash} an open-source, scalable automated machine learning (AutoML) package {\textemdash} and enhance its capability to generate MPNN neural architectures for molecular property prediction. We evaluate the efficacy of our approach on quantum mechanics and physical chemistry datasets from the MoleculeNet benchmark \cite{wu2018moleculenet}. The contributions of the paper are as follows.

\begin{enumerate}
    \item We develop an NAS approach to generate stacked MPNN architecture to predict the molecular properties of small molecules.
    \item We show that automatically obtained stacked MPNNs compare favorably with manually designed networks to predict molecular properties on three quantum mechanics and three physical chemistry datasets from the MoleculeNet benchmark.
    \item DeepHyper has been utilized for automated discovery of fully connected neural networks on tabular data \cite{balaprakash2018deephyper} and long short-term memory (LSTM) on time series data \cite{maulik2020recurrent}. In this study, we enhance  DeepHyper's capabilities for discovering stacked MPNN architectures. 
\end{enumerate}

\section{MoleculeNet Benchmark}

We use the small molecule datasets provided by MoleculeNet \cite{wu2018moleculenet} that include two groups of datasets.

\subsection{Quantum Physics}
The quantum physics group comprises QM7, QM8, and QM9 datasets. The QM7 dataset is a subset of the GDB-13 database \cite{blum}, which contains about 1 billion organic molecules with up to seven ``heavy" atoms (C, N, O, S). The QM7 dataset includes 7,160 molecules and their corresponding atomization energies. The QM8 dataset is a subset of the GDB-17 database \cite{ruddigkeit2012enumeration}. The dataset contains four excited-state properties of 21,786 molecules with up to eight heavy atoms collected by using three methods, including time-dependent DFTs and second-order approximate coupled-cluster. The QM9 dataset provides twelve properties, including geometric, energetic, electronic, and thermodynamic properties, for a subset of the GDB-17 database \cite{ruddigkeit2012enumeration}. The dataset is made up of 133,865 molecules with up to nine heavy atoms, all modeled using DFT. 

\subsection{Physical Chemistry}
The physical chemistry groups consists of ESOL, Free Solvation Database (FreeSolv), and Lipophilicity datasets. The ESOL dataset comprises water solubility data for 1,128 molecules \cite{delaney2004esol}. The  FreeSolv dataset consists of experimental and calculated hydration-free energy of 642 small molecules in water \cite{mobley2014freesolv}. The Lipophilicity dataset, chosen from the ChEMBL database  \cite{davies2015chembl}, provides experimental results of the octanol/water distribution coefficient (important feature for drug design) for 4,200 molecules.


\subsection{Molecule Featurization}
MoleculeNet provides multiple methods to represent molecules. In our study, we use the Weave featurizer that encodes both local chemical environment and connectivity between atoms. Specifically, node (atomic) features are in vector format, while the connectivity is represented by a list of edge (pair) features. A node feature vector $\bh \in \mathbb{R}^{75}$ contains information on atom type, hybridization type, and valence structure. An edge vector $\be \in \mathbb{R}^{14}$ encodes bond types, graph distance, and ring information.

\section{Neural Architecture Search}
\label{sec:search_space}

The NAS approach that we propose comprises three components: (i) a search space that defines a set of stacked MPNN feasible architectures; (ii) a search method that will explore the search space to find the best architecture; and (iii) an evaluation method that computes the accuracy of an MPNN architecture sampled using the search method. 

\begin{figure}[tb]
    \centering
    \includegraphics[width=0.8\columnwidth]{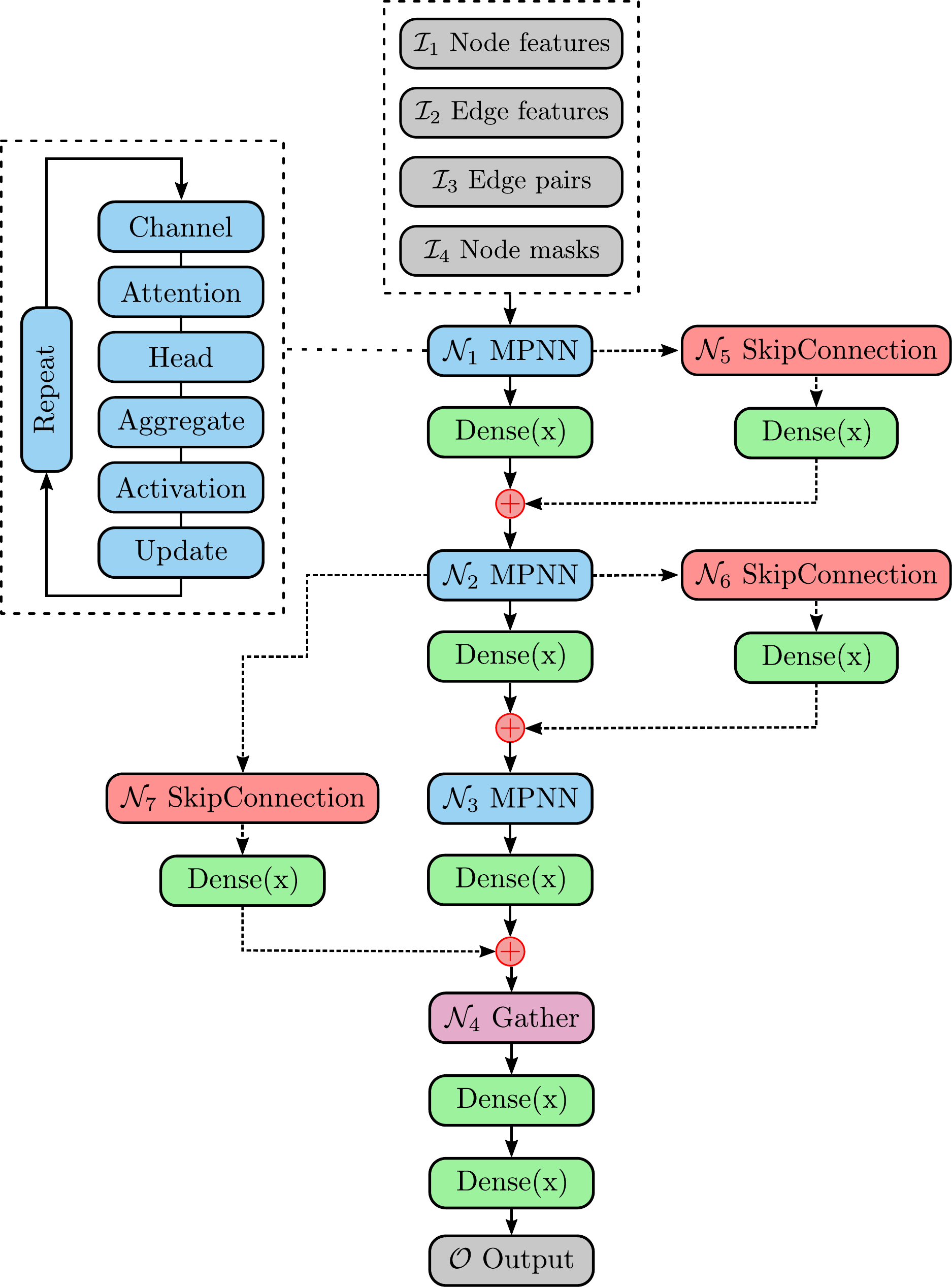}
    \caption{Example stacked MPNN search space with three MPNN variable nodes in blue: $\mathcal{N}_1$, $\mathcal{N}_2$, and $\mathcal{N}_3$. The skip-connection variable nodes are $\mathcal{N}_5$, $\mathcal{N}_6$, and $\mathcal{N}_7$. Dotted lines represent possible skip connections. The gather variable node is $\mathcal{N}_4$. The inputs to the networks are node features $\mathcal{I}_1$, edge features $\mathcal{I}_2$, edge pairs $\mathcal{I}_3$, and node masks $\mathcal{I}_4$. After two constant dense nodes with 32 hidden units, the output node is $\mathcal{O}$. The variables in the MPNN node include the number of channels, the attention mechanisms, the number of attention heads, the aggregation methods, the activation functions, the update functions, and the number of repetitions.}
    \label{fig:search_space}
\end{figure}

\subsection{Stacked MPNN Search Space}

We define the MPNN search space as a directed acyclic graph. An example stacked MPNN search space is shown in Fig.~\ref{fig:search_space}. The input and output nodes are fixed and denoted as $\mathcal{I}$ and $\mathcal{O}$, respectively. All other nodes $\mathcal{N}$ are intermediate nodes, each containing a list of possible operations. Intermediate nodes are made up of two categories: constant node (single operation) and variable node (multiple operations). For a given variable node, an index is assigned for each operation. An architecture from the space can be defined by using a vector $\bp \in \mathbb{Z}^n$, where $n$ is the number of variable nodes. Each entry $\bp_i$ is an index chosen from a set of possible index values for the variable node $i$. The stacked MPNN search space is composed of MPNN, skip-connection, and gather variable nodes, described below. 



\subsubsection{Input node}
As shown in Fig.~\ref{fig:search_space}, the inputs for any given network include node features, edge features, edge pairs, and node masks. 
For a given molecule dataset, $N$ and $E$ are the maximum number of nodes (atoms) and edges (bonds), respectively. We use zero padding to create the node feature matrix to $\bH \in \mathbb{R}^{N \times F_n}$ and the edge feature matrix to $\bE \in \mathbb{R}^{E \times F_e}$, where $F_n$ and $F_e$ are the numbers of node features and edge features, respectively. We also have an edge pair matrix $\bP \in \mathbb{Z}^{E \times 2}$, where each row contains the indices of two nodes connected by a given edge. Because molecules have a varied number of atoms (nodes), the node mask vector $\bm \in \mathbb{Z}^N$ is used to screen out the non-existent node features that are added by zero padding. An existent node has $\bm_i = 1$, and a nonexistent node has $\bm_i = 0$.

\subsubsection{MPNN node}
Each MPNN node runs internally for $T$ time steps to update the hidden feature of each node. An MPNN node contains a message function $M_t$ and an update function $U_t$,  defined as follows:
\begin{align}
    \bm_v^{t+1} &= \text{Agg}_{w\in\mathcal{N}\rb{v}}M_t \rb{\bh_v^t, \bh_w^t, \be_{vw}} \label{eqn:mpnn-1}\\
    \bh_v^{t+1} &= U_t \rb{\bh_v^t, \bm_v^{t+1}} \label{eqn:mpnn-2}.
\end{align}
To update the hidden feature of a node $v$, the message function  $M_t$ at step $t$ takes as inputs the node $v$ feature $\bh_v^t$, the neighboring node feature $\bh_w^t$ for $w \in \mathcal{N}\rb{v}$, and the edge feature $\be_{vw}$ between node $v$ and $w$. The output of the message function $M_t$ contains a list of message vectors from neighboring nodes. The aggregate function Agg collects the message vectors and generates the intermediate hidden feature $\bm_{v}^{t+1}$. The aggregate function Agg is 
one of mean, summation, or max pooling. The update function $U_t$ at step $t$ combines the node feature $\bh_v^t$ and the intermediate hidden feature $\bm_v^{t+1}$ to create the new hidden feature of step $t+1$ $\bh_v^{t+1}$. The detailed structures of the message function $M_t$ and update function $U_t$ are as follows.
\begin{align}
    M_t \rb{\bh_v^t, \bh_w^t, \be_{vw}} &= \alpha_{vw}\text{MLP}\rb{\be_{vw}}\bh_w^t \label{eqn:message-1}\\
    U_t \rb{\bh_v^t, \bm_v^{t+1}} &= 
    \begin{cases}
    \text{GRU}\rb{\bh_v^t, \bm_v^{t+1}} \label{eqn:update-1}\\
    \text{MLP}\rb{\bh_v^t, \bm_v^{t+1}}
    \end{cases}.
\end{align}
Typically, the message function has a multilayer perceptron (MLP or edge network) to handle the edge feature $\be_{vw}$. The processed edge feature is multiplied with $\bh_w^t$ to yield a message from node $w$ to $v$. In this case, the processed edge feature MLP$\rb{\be_{vw}}$ can be viewed as a weight for $\bh_w^t$. Borrowing the idea of node attention, we add an attention coefficient $\alpha_{vw}$ to further modify the weight of $\bh_w^t$. The attention coefficient is a function of $\bh_v^t$ and $\bh_w^t$. The update function $U_t$ can be either  a gated recurrent unit (GRU) or an MLP. To further elucidate the design of the MPNN node, we divide it  into the following five categories of operations.
\begin{enumerate}
    \item \textbf{State dimension}: After running $T$ times, the MPNN node maps the input node feature to a $d$-dimensional vector. The choice of state dimension $d$ is important for final prediction. To reduce the number of parameters and increase the generalizability, the set of state dimensions is set to $\{4, 8, 16, 32\}$.
    \item \textbf{Attention function}: Although the information passing weight between nodes is governed by the edge feature in traditional MPNNs, the attention mechanism helps focus on the most relevant neighboring nodes to improve information aggregation. Following NAS frameworks in \cite{zhou2019auto, gao2019graphnas}, the attention functions used to calculate coefficient $\alpha_{vw}$ are shown in Table \ref{tab:attention}. For constant attention, $\alpha_{vw}$ is always 1. For GCN attention, $\alpha_{vw}$ is $\frac{1}{\sqrt{|\mathcal{N}\rb{v}| \cdot |\mathcal{N}\rb{w}|}}$, where $|\mathcal{N}\rb{v}|$ and $|\mathcal{N}\rb{w}|$ represent the number of neighboring nodes for node $v$ and $w$, respectively. For GAT, $\alpha_{vw}$ is $\text{LeakyReLU}\rb{\ba \rb{\bW\bh_v||\bW\bh_w}}$, where $\ba$ is a trainable vector, $\bW$ is a trainable weight matrix, $||$ denotes concatenation, and $\bh_v$ and $\bh_w$ are the node hidden feature vectors for nodes $v$ and $w$, respectively. By adding $\alpha_{vw}$ with $\alpha_{wv}$ from GAT, we obtain the attention coefficient for SYM-GAT. For COS attention, $\alpha_{vw}$ is $\ba\rb{\bW\bh_v||\bW\bh_w}$. For linear attention, $\alpha_{vw}$ is tanh$\rb{\ba_l\bW\bh_v+\ba_r\bW\bh_w}$, where $\ba_l$ and $\ba_r$ are two trainable vectors. For gen-linear attention, $\alpha_{vw}$ is $\bW_G\text{tanh}\rb{\bW\bh_v+\bW\bh_w}$, where $\bW_G$ is the trainable matrix. 
    \item \textbf{Attention head}: Multihead attention could be useful to stabilize the learning process \cite{velivckovic2017graph, lee2019attention}. We select the number of heads from the set of $\{1, 2, 4, 6\}$.
    \item \textbf{Aggregate function}: Aggregation function is critical to capture the neighborhood structures for extracting node representation \cite{xu2018powerful}. The aggregation functions are selected from the set of $\{$mean, summation, max-pooling$\}$.
    \item \textbf{Activation function}: Following NAS frameworks in \cite{zhou2019auto, gao2019graphnas}, the  possible activation functions used in the MPNN are $\{$Sigmoid, Tanh, ReLU, Linear, Softplus, LeakyReLU, ReLU6,  ELU$\}$.
    \item \textbf{Update function}: Node features $\bh_v^t$ and intermediate hidden feature $\bm_v^{t+1}$ are combined and propagated by an update function to generate the new feature $\bh_v^{t+1}$. The update functions we use are $\{$GRU, MLP$\}$.
\end{enumerate}

\begin{table}[htbp]
\centering
\caption{Set of attention functions, where $||$ denotes the concatenation; $\ba, \ba_{l}, and \ba_{r}$ are the trainable vectors; and $\bW_G$ is the trainable matrix}
\label{tab:attention}
\begin{tabular}{@{}l|l@{}}
\toprule
Attention Mechanism & Equations \\ 
\midrule
Constant & 1 \\ 
\midrule
GCN & $\frac{1}{\sqrt{|\mathcal{N}\rb{v}||\mathcal{N}\rb{w}|}}$ \\
\midrule
GAT & LeakyReLU$\rb{\ba \rb{\bW\bh_v||\bW\bh_w}}$ \\ 
\midrule
SYM-GAT & $\alpha_{vw}+\alpha_{wv}$ based on GAT \\ 
\midrule
COS & $\ba\rb{\bW\bh_v||\bW\bh_w}$ \\
\midrule
Linear & tanh$\rb{\ba_l\bW\bh_v+\ba_r\bW\bh_w}$ \\
\midrule
Gen-linear & $\bW_G\text{tanh}\rb{\bW\bh_v+\bW\bh_w}$ \\
\bottomrule
\end{tabular}
\end{table}


The skip-connection node is a special case of variable node. Given three nodes $\mathcal{N}_{i-1}$, $\mathcal{N}_i$, and $\mathcal{N}_{i+1}$ in a sequence, the skip-connection allows the connection between $\mathcal{N}_{i-1}$ and $\mathcal{N}_{i+1}$. The skip-connection includes two operations: identity for skip-connection and empty for no skip-connection. In a skip-connection operation, the tensor output from $\mathcal{N}_{i-1}$ is processed by a dense layer and a summation operator. Because the output from variable nodes can have different shapes, the dense layer projects the incoming tensor to a right shape for summation. The summation operator adds the output from $\mathcal{N}_{i-1}$ and $\mathcal{N}_i$ and passes the result to $\mathcal{N}_{i+1}$. The skip-connection operation can be applied to any length of node sequences. In our study, we limit the skip-connection to cross at a maximum of three nodes. For example, $\mathcal{N}_{i-1}$ can be added with $\mathcal{N}_{i+2}$ to be passed to $\mathcal{N}_{i+3}$.

\subsubsection{Gather variable node}
The gather node contains eleven operations belonging to five categories. Let $\bH \in \mathbb{R}^{N \times F}$ be the node feature input to the gather node, where $N$ is the number of nodes and $F$ is the number of hidden features. Following the gather operations in the Spektral \cite{grattarola2020graph} GNN package, the graph operations in stacked MPNN can be divided into five categories.
\begin{enumerate}
    \item \textbf{Global pool}: Pools a graph by computing the sum, mean, or  maximum of its node features. The output has a shape of $\mathbb{R}^{N}$.
    \item \textbf{Global gather}: Calculates the sum, mean, or  maximum of a feature for all the nodes. The output has a shape of $\mathbb{R}^{F}$.
    \item \textbf{Global attention pool}:   Computes \cite{li2015gated}  the output $\bH_{out} \in \mathbb{R}^{F'}$ as $\bH_{out} = \sum_{i=1}^{N}\rb{\sigma \rb{\bH\bW_1 + \bb_1} \odot \rb{\bH\bW_2 + \bb_2}}_i$, where $\sigma$ is the sigmoid activation function; $\bW_1, \bW_2$ are trainable weights; and $\bb_1, \bb_2$ are biases. The output dimension $F'$ is selected to be in $\{16, 32, 64\}$.
    \item \textbf{Global attention sum pool}: Pools a graph by learning attention coefficients to sum node features. The operation can be defined as $\bH_{out} = \sum_{i=1}^{N} \alpha_i \cdot \bH_i$, where $\alpha = \text{softmax}\rb{\bH\ba}$ and $\boldsymbol{\alpha} \in \mathbb{R}^F$ is a trainable vector. The softmax activation is applied across nodes.
    \item \textbf{Flatten}: Flattens $\bH$ to a 1D vector.
\end{enumerate}

\subsection{Search Methods}

To find a high-performing MPNN from the search space, we adopt regularized evolution (RE) \cite{real2019regularized}, an asynchronous search method implemented and available in the DeepHyper package. RE searches for new architectures by applying a mutation to existing models within a population. It starts the search with a population of $P$ random architectures, evaluates them, and records the validation loss from each individual. Following the initialization, it samples $S$ random architectures uniformly from the population with replacement. The architecture with the lowest validation loss within the sample is selected as a parent. A mutation is performed on the parent, and a new child architecture is constructed. A mutation corresponds to choosing a different operation for a variable node in the search space. Specifically, RE randomly samples a variable node with a random operation that excludes the current value. The child is trained, and the validation loss is recorded. Consequently, the child is added to the population by replacing the oldest architecture in the population. Over multiple cycles, architectures with lower validation loss are retained in the population via repeated sampling and mutation. The sampling and mutation are computationally inexpensive and can be performed quickly. When RE finishes an evaluation, a new architecture for training is obtained by the mutation on the previously evaluated architecture (stored throughout the duration of the experiment in memory).
The key advantage of RE stems from its scalability. It can leverage multiple compute nodes to evaluate architectures in parallel, which results in faster convergence to high-performing architectures. It has been shown that RE, because of its minimal algorithmic overhead and synchronization, outperforms reinforcement learning methods for NAS \cite{real2019regularized,maulik2020recurrent}.


As a baseline, we also include random search (RS) that explores the search space by randomly assigning an operation to each variable node. Specifically, RS samples a random vector of indices to represent the architecture and trains the architecture independently from other nodes. Although the search is parallel, the discovered architectures typically do not improve over time because of the lack of a feedback loop. However, comparison with RS is critical for the new dataset to quantify the effectiveness of the RE search method.


\subsection{Evaluation Strategy}

Each evaluation consists of training a generated network and computing the accuracy metric (reward) on the validation data. The evaluation uses a single node (no multinode data-parallel training). Following the implementation of MoleculeNet, the QM7, QM8, and QM9 datasets are evaluated by mean absolute error (MAE); ESOL, FreeSolv, and Lipophilicity are evaluated by the root mean square error (RMSE). RE maximizes the reward; therefore, negative MAE and RMSE are given the reward. To enable rapid  exploration of the search space, we use a subset of the training dataset and fewer epochs for training. After the search is over, the best architecture (selected based on the reward) is retrained with the full training data and a larger number of epochs. 

\section{Experiments}

We used Bebop, a 1,024-node cluster at Argonne's Laboratory Computing Resource Center. Bebop has two types of nodes: Broadwell and Knights Landing. We used Broadwell nodes for the experiments. Each node is a 36-core Intel Xeon E5-2695v4 processor with 128 GB of DDR4 memory. 
The compute nodes are interconnected by Omni-Path Fabric. The software environment that we used consists of Python 3.7.6, TensorFlow 1.14 \cite{tensorflow2015-whitepaper}, DeepHyper 0.1.11, DeepChem 2.4, and Spektral 0.1.2. DeepHyper NAS API utilized TensorFlow Keras. 

For the stacked MPNN search space, we used three MPNN variable nodes ($m=3$), resulting in the creation of six skip-connection variable nodes. We used one gather variable node and two constant dense nodes with 32 hidden units between gather node and prediction. 
The total number of architectures in the search space is 23,626,761,124,184,064 ($\approx 2\times 10^{16}$).

As shown in Fig.~\ref{fig:search_space}, the inputs are node features, edge features, edge pairs, and node masks. We use zero padding to create the node feature matrix to $\bH \in \mathbb{R}^{N \times 75}$ and the edge feature matrix to $\bE \in \mathbb{R}^{E \times 14}$, where $N$ is the number of nodes, E is the number of edges, 75 is the number of node features, and 14 is the number of edge features. Our edge pair matrix $\bP \in \mathbb{Z}^{E \times 2}$ contains the source and target node indices of the edges. The graph can be traversed from the source node to the target node. We treat molecules as undirected graphs, where edges are bidirectional. Specifically, the graph can be traversed from node $i$ to node $j$, as well as from node $j$ to node $i$. We also add self-loop, which is an edge that connects a node with itself. The node mask vector $\bm \in \mathbb{Z}^N$ has entries of 1 when nodes exist and 0 when nodes do not exist.
The optimizer for training was ADAM \cite{kingma2014adam}.


For the split of training, validation, and test data, we followed the MoleculeNet implementation and used the given stratified splitter to split the QM7 dataset and the random splitter to split other datasets using fixed random seeds. The training, validation, and test split ratio is 8:1:1. Specifically, QM7 has 5,728 training, 716 validation, and 716 test data. The maximum number of nodes $N$ is 7, and the maximum number of edges $E$ is 9. QM8 has 17,428 training, 2,179 validation, and 2,179 test data with $N=9$ and $E=14$. QM9 has 107,107 training, 13,389 validation, and 13,389 test data with $N=9$ and $E=16$. ESOL has 902 training, 113 validation, and 113 test data with $N=55$ and $E=68$. FreeSolv has 512 training, 65 validation, and 65 test data with $N=24$ and $E=25$. Lipophilicity has 3,360 training, 420 validation, and 420 test data with $N=115$ and $E=236$.

RE was configured to run with a population size ($P$) of  100 and a sample size ($S$) of 10 to process the mutation of architectures. RS evaluated a random architecture on each node. Since both RE and RS were asynchronous, all nodes were workers that could evaluate architectures independently without any master.

MoleculeNet has various benchmark models, including random forest (RF), graph convolution (GC), deep tensor network (DTNN), message-passing neural network (MPNN), and Weave network. Only MPNN and Weave network use Weave featurizer to represent molecule data with both atom (node) and node (edge) information. Because stacked MPNN uses the same Weave featurizer to generate data, we compared it with the better MPNN and Weave network from MoleculeNet.

For the evaluation, we limited the training time for each architecture to less than 10 minutes. QM8 and Lipophilicity training data were halved. QM9 training data were shrunk to one-tenth. Other data sets had full training data. QM7, QM8, and QM9 were trained for 50, 40, and 50 epochs, respectively. ESOL, FreeSolv, and Lipophilicity were trained for 80, 200, and 20 epochs, respectively. For post-training, we selected the architecture discovered to have the lowest validation loss and trained it from scratch for 200 epochs using the full dataset. Following the implementation of MoleculeNet, three independent runs with different random seeds were performed. The three fixed randoms seeds were used to split the dataset into training, validation, and testing sets. The results were the average accuracy metrics of three runs with standard deviations as errors. MoleculeNet provided various data representations for molecules. Each MoleculeNet featurizer could generate a unique data representation. Both stacked MPNN and MoleculeNet GNN benchmark results were based on the same data representation generated by the Weave featurizer, except for the QM7 dataset, for which the MoleculeNet benchmark used the GCN featurizer to generate molecule representations without edge features.

\subsection{Neural Architecture Search Using Regularized Evolution}
Here, we show that RE discovers high-performing architectures within short computation times and that these architectures outperform the manually designed baseline GNN architectures from MoleculeNet on the six datasets.

\begin{figure}[tb]
\centering
\includegraphics[width=0.9\columnwidth]{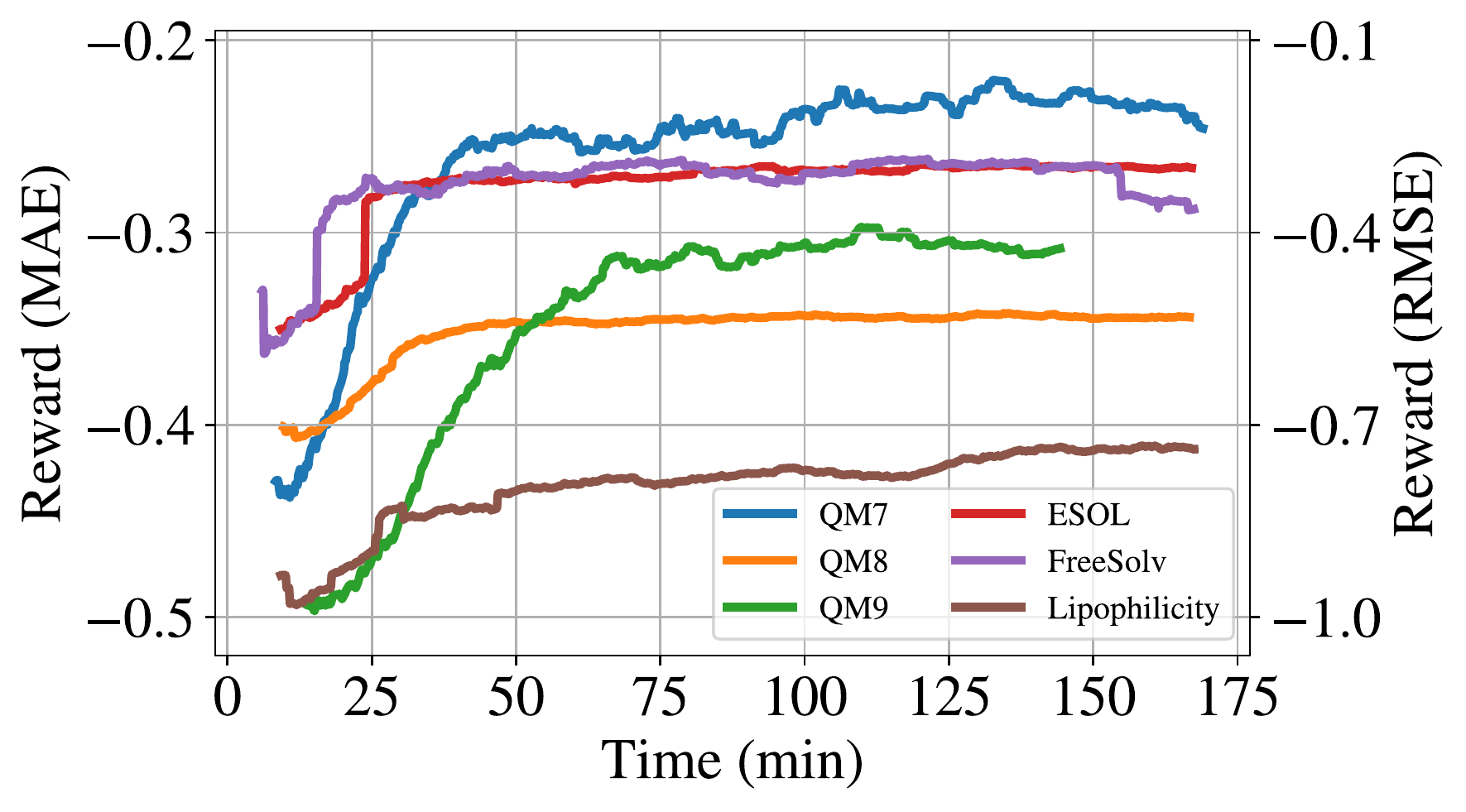}
\caption{Search trajectories for all datasets. The reward for QM7, QM8, and QM9 is the negative validation MAE. The reward for ESOL, FreeSolv, and Lipophilicity is the negative validation RMSE.}
\label{fig:search_trajectory}
\end{figure}

We ran the RE search method on 30 nodes for 3 hours of wall-clock time.  Figure~\ref{fig:search_trajectory} shows the search trajectory of reward (negative MAE/RMSE) with respect to wall-clock time. We calculated the search trajectory by using a running average of window size 100 on the reward obtained by the architectures found during the search. For the quantum mechanics datasets (QM7, QM8, and QM9), RE converged after 50, 40, and 75 minutes, respectively. The longer convergence time on QM9 can be attributed to the large dataset size, which made each individual training slower. For the physical chemistry datasets (ESOL, FreeSolv, and Lipophilicity), RE  converged after 25, 20, and 75 minutes, respectively. Because ESOL and FreeSolv were small datasets, they were expected to have shorter convergence times. The Lipophilicity dataset had more samples, and each sample was larger, thus msking the convergence slower.



\begin{figure}[tb]
\centering
\begin{subfigure}{0.45\columnwidth}
  \centering
  \includegraphics[width=\columnwidth]{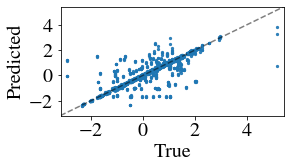}  
  \caption{QM7}
  \label{fig:qm7_parity}
\end{subfigure}
\begin{subfigure}{0.45\columnwidth}
  \centering
  \includegraphics[width=\columnwidth]{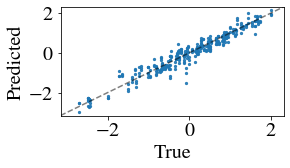}  
  \caption{ESOL}
  \label{fig:esol_parity}
\end{subfigure}

\begin{subfigure}{0.45\columnwidth}
  \centering
  \includegraphics[width=\columnwidth]{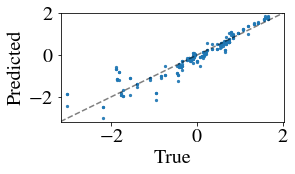}  
  \caption{FreeSolv}
  \label{fig:freesolv_parity}
\end{subfigure}
\begin{subfigure}{0.45\columnwidth}
  \centering
  \includegraphics[width=\columnwidth]{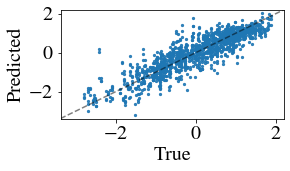}  
  \caption{Lipophilicity}
  \label{fig:lipo_parity}
\end{subfigure}
\caption{Parity plots of the single-class regression test datasets.}
\label{fig:parity}
\end{figure}

The single-class regression results for each test dataset are shown in Table~\ref{tab:comparison}. The MAE of stacked MPNN on QM7 is 48.0$\pm$0.7, which is a 38\% drop from the MoleculeNet MAE of 77.9$\pm$2.1. The RMSEs of stacked MPNN on ESOL and Lipophilicity are 0.54$\pm$0.01 and 0.598$\pm$0.043, respectively, which are 7\% and 16\% lower than the MoleculeNet results of 0.58$\pm$0.03 and 0.715$\pm$0.035. MoleculeNet has a lower RMSE of 1.15$\pm$0.12 on FreeSolv compared with 1.21$\pm$0.03 on stacked MPNN. FreeSolv is a small dataset with 64 molecules in each test set. The difference of random seeds used in MoleculeNet and the over-fitting problem brought on by a small dataset can lead to a worse result for stacked MPNN. The regression parity plots for single-class regression datasets are shown in Fig.~\ref{fig:parity}. All test set prediction points aligned close to the diagonal line,  indicating that the predicted values of stacked MPNN are close to the true values.

QM8 and QM9 are multiclass regression datasets. As shown in Table~\ref{tab:compare-qm8}, stacked MPNN has a smaller MAE for each feature in the multiclass regression of QM8 compared with MoleculeNet benchmarks. Moreover, Table~\ref{tab:compare-qm8} illustrates that for most of the features, stacked MPNN outmatches MoleculeNet GNN except for mu and R2. In particular, the R2 value is at least a magnitude larger than other features. The suboptimal loss value at which the training was stuck could lead to the inconsistency in the R2 value \cite{chen2019alchemy}.

\begin{table}[tb]
\centering
\caption{Single-class regression test set performance comparison}
\label{tab:comparison}
\begin{tabular}{c|cc}
\toprule
Data Set & Stacked MPNN & MoleculeNet GNN \\
\midrule
QM7 (MAE) & \bf{48.0$\pm$0.7} & 77.9$\pm$2.1 \\
ESOL (RMSE) & \bf{0.54$\pm$0.01} & 0.58$\pm$0.03 \\
FreeSolv (RMSE) & 1.21$\pm$0.03 & \bf{1.15$\pm$0.12} \\
Lipophilicity (RMSE) & \bf{0.598$\pm$0.043} & 0.715$\pm$0.035 \\
\bottomrule
\end{tabular}
\end{table}

\begin{table}[tb]
\centering
\caption{QM8 and QM9 test set performance comparison (MAE)}
\label{tab:compare-qm8}
\begin{adjustbox}{width=\columnwidth}
\begin{tabular}{c|cc|c|cc}
\toprule
\multicolumn{1}{c|}{QM8} & \multicolumn{1}{c}{\begin{tabular}[c]{@{}c@{}}Stacked \\ MPNN\end{tabular}} & \multicolumn{1}{c|}{\begin{tabular}[c]{@{}c@{}}Molecule\\ Net GNN\end{tabular}} & \multicolumn{1}{c|}{QM9} & \multicolumn{1}{c}{\begin{tabular}[c]{@{}c@{}}Stacked \\ MPNN\end{tabular}} & \multicolumn{1}{c}{\begin{tabular}[c]{@{}c@{}}Molecule\\ Net GNN\end{tabular}} \\
\midrule
E1-CC2 & \bf{0.0068$\pm$0.0003} & 0.0084 & mu & 0.564$\pm$0.003 & \bf{0.358} \\
E2-CC2 & \bf{0.0079$\pm$0.0003} & 0.0091 & alpha & \bf{0.69$\pm$0.01} & 0.89 \\
f1-CC2 & \bf{0.0129$\pm$0.0002} & 0.0151 & HOMO & \bf{0.00560$\pm$0.00004} & 0.00541 \\
f2-CC2 & \bf{0.0291$\pm$0.0005} & 0.0314 & LUMO & \bf{0.00602$\pm$0.00002} & 0.00623 \\
E1-PBE0 & \bf{0.0064$\pm$0.0001} & 0.0083 & gap & \bf{0.0080$\pm$0.0000} & 0.0082 \\
E2-PBE0 & \bf{0.0072$\pm$0.0001} & 0.0086 & gap & \bf{0.0080$\pm$0.0000} & 0.0082 \\
f1-PBE0 & \bf{0.0104$\pm$0.0002} & 0.0123 & gap & \bf{0.0080$\pm$0.0000} & 0.0082 \\
f2-PBE0 & \bf{0.0216$\pm$0.0005} & 0.0236 & gap & \bf{0.0080$\pm$0.0000} & 0.0082 \\
E1-CAM & \bf{0.0063$\pm$0.0001} & 0.0079 & R2 & 41.3$\pm$0.6 & \bf{28.5} \\
E2-CAM & \bf{0.0069$\pm$0.0002} & 0.0082 & ZPVE & \bf{0.00130$\pm$0.00014} & 0.00216 \\
f1-CAM & \bf{0.0117$\pm$0.0002} & 0.0134 & ZPVE & \bf{0.00130$\pm$0.00014} & 0.00216 \\
f2-CAM & \bf{0.0236$\pm$0.0001} & 0.0258 & U0 & \bf{0.65$\pm$0.06} & 2.05 \\
& & & U & \bf{0.62$\pm$0.05} & 2.00 \\
& & & H & \bf{0.68$\pm$0.11} & 2.02 \\
& & & G & \bf{0.66$\pm$0.05} & 2.02 \\
& & & Cv & \bf{0.35$\pm$0.01} & 0.42 \\
\bottomrule
\end{tabular}
\end{adjustbox}
\end{table}



\subsection{Search Space Analysis}


Using the architectures obtained from the RE runs, we analyzed the relative importance of each operation in the NAS search space for a given dataset.
We assigned a vector $a$ for each architecture explored during the search. Each entry $a_i$ is either -1 or 1, where -1 denotes the absence of a certain operation and 1 denotes the existence of the operation. For example, the first MPNN node can have four choices of state dimension: 4, 8, 16, and 32. In a given architecture, if the state dimension of the first MPNN node is 32, the entry of dim(32)[cell1] is 1, whereas other choices such as dim(4)[cell1], dim(8)[cell1], and dim(16)[cell1] are absent and therefore -1.

We trained a random forest model with 100 trees to predict the search reward from the operation vector $a$. A decision tree with $L$ leaves divides the feature space into $L$ regions $R_l$ for $l = 1, \cdots, L$. The prediction function of a tree is defined as $f\rb{a} = \sum_{l=1}^L c_l I\rb{a, R_l} \label{eqn:tree-1}$, where $L$ is the number of leaves in the tree, $R_L$ is a region in the search space that corresponds to leaf $l$, $c_l$ is a constant associated with region $l$, and $I$ is an indicator function: specifically, $I\rb{a, R_l} = 1$ if $a \in R_l$, and $0$ otherwise. The constant $c_l$ is determined during the training of the tree; $c_l$ represents the mean of the response variables of samples that belong to region $R_l$.

\begin{figure}[tb]
\centering
\includegraphics[width=0.8\columnwidth]{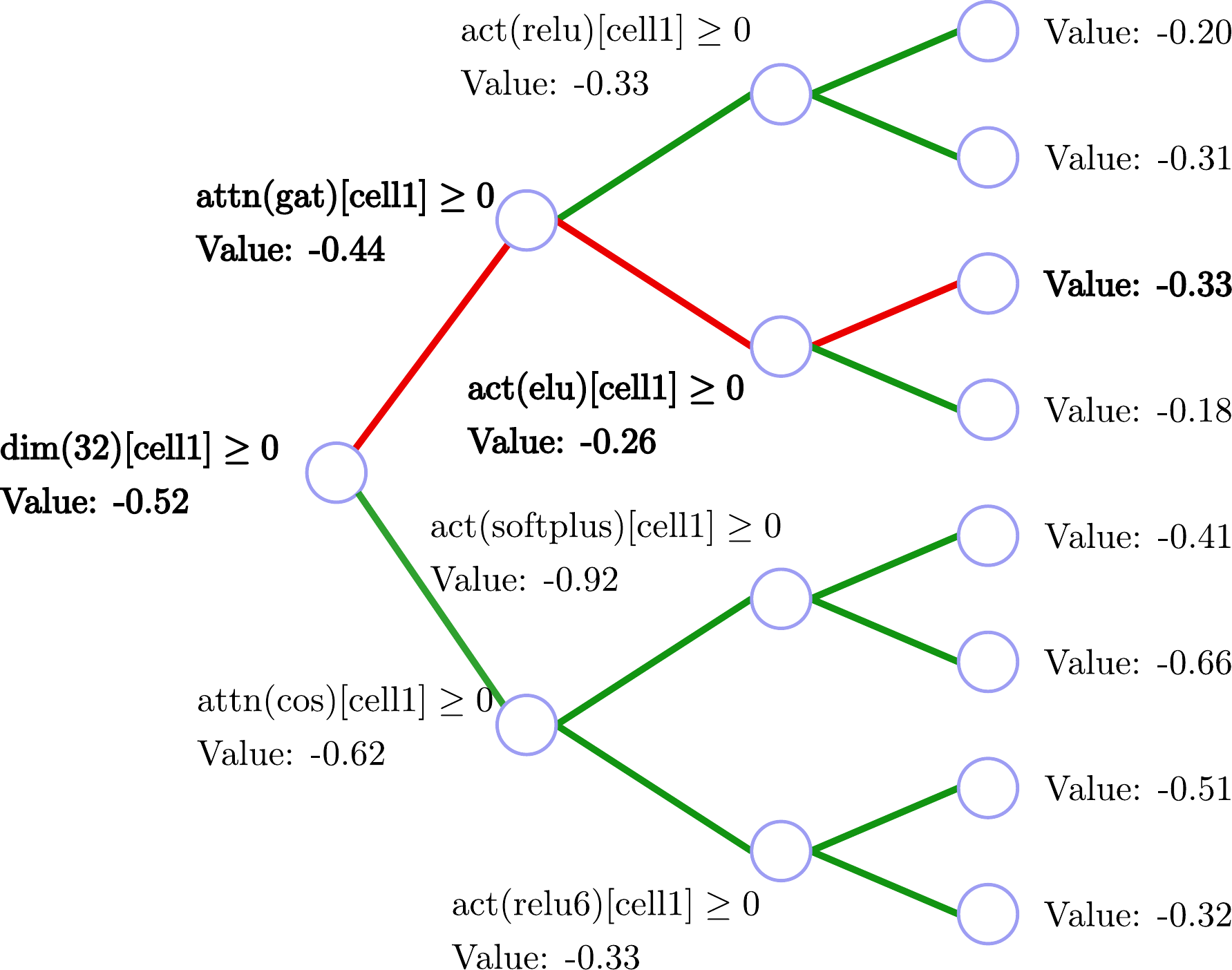}  
\caption{Example decision tree that represents a search space.}
\label{fig:decision-tree}
\end{figure}

Following the decision path, each decision is governed by an operation that either adds or subtracts from the value of the parent node. The tree interpreter \cite{Divingintodata} therefore defines the regression tree prediction function as $f\rb{a} = c_{full} + \sum_{k=1}^K contrib\rb{a, k}\label{eqn:tree-3}$, where $K$ is the number of operations and $c_{full}$ is the value (i.e., the mean given by the topmost region that covers the entire training set) at the root node. For convenience, the root node value is denoted as \textit{bias}, and  $contrib\rb{a,k}$ is the contribution of $k$th operation in the operation vector $a$.

Because the prediction of a forest is the average of the predictions of its trees, the prediction function of a random forest can be defined as $F\rb{a} = \frac{1}{J} \sum_{j=1}^J f_{j}\rb{a}\label{eqn:tree-4}$, where $J$ is the number of trees in the forest. Following the decision path notation, we can break down the prediction function to $F\rb{a} = \frac{1}{J}c_{j_{full}} + \sum_{k=1}^K \rb{\frac{1}{J} contrib_j\rb{a,k}}$.

In Fig.~\ref{fig:decision-tree} we show an example decision tree. The tree has a condition on each internal node and a value associated with each leaf (i.e., the value to be predicted). In addition, the value at each node is the average of the response variables in that region. Specifically, the root node's value -0.52 represents the bias of the training set. Because the operation vector $a$ is binary, each node decides whether an operation exists or not. Following the red path in Fig.~\ref{fig:decision-tree}, the prediction value -0.25 can be calculated as -0.25 = -0.52(training set mean) + 0.08(gain from the presence of dim(32)[cell1]) + 0.18(gain from the absence of attn(gat)[cell1]) - 0.07(loss from the presence of act(elu)[cell1]).

\begin{figure*}[h!]
\centering
\begin{subfigure}{0.6\columnwidth}
  \centering
  \includegraphics[width=\columnwidth]{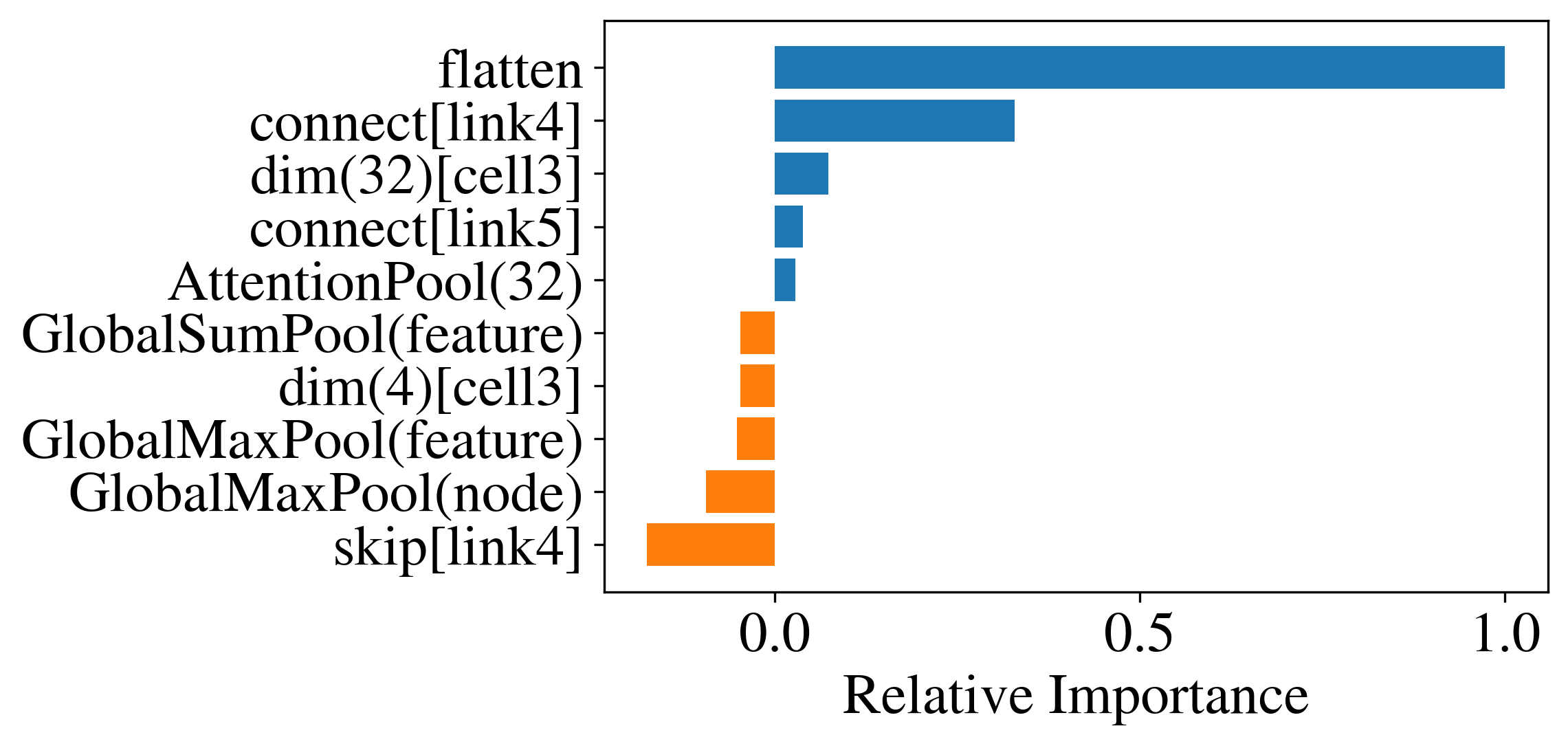}
  \caption{QM7}
  \label{fig:qm7_importance}
\end{subfigure}
\begin{subfigure}{0.6\columnwidth}
  \centering
  \includegraphics[width=\columnwidth]{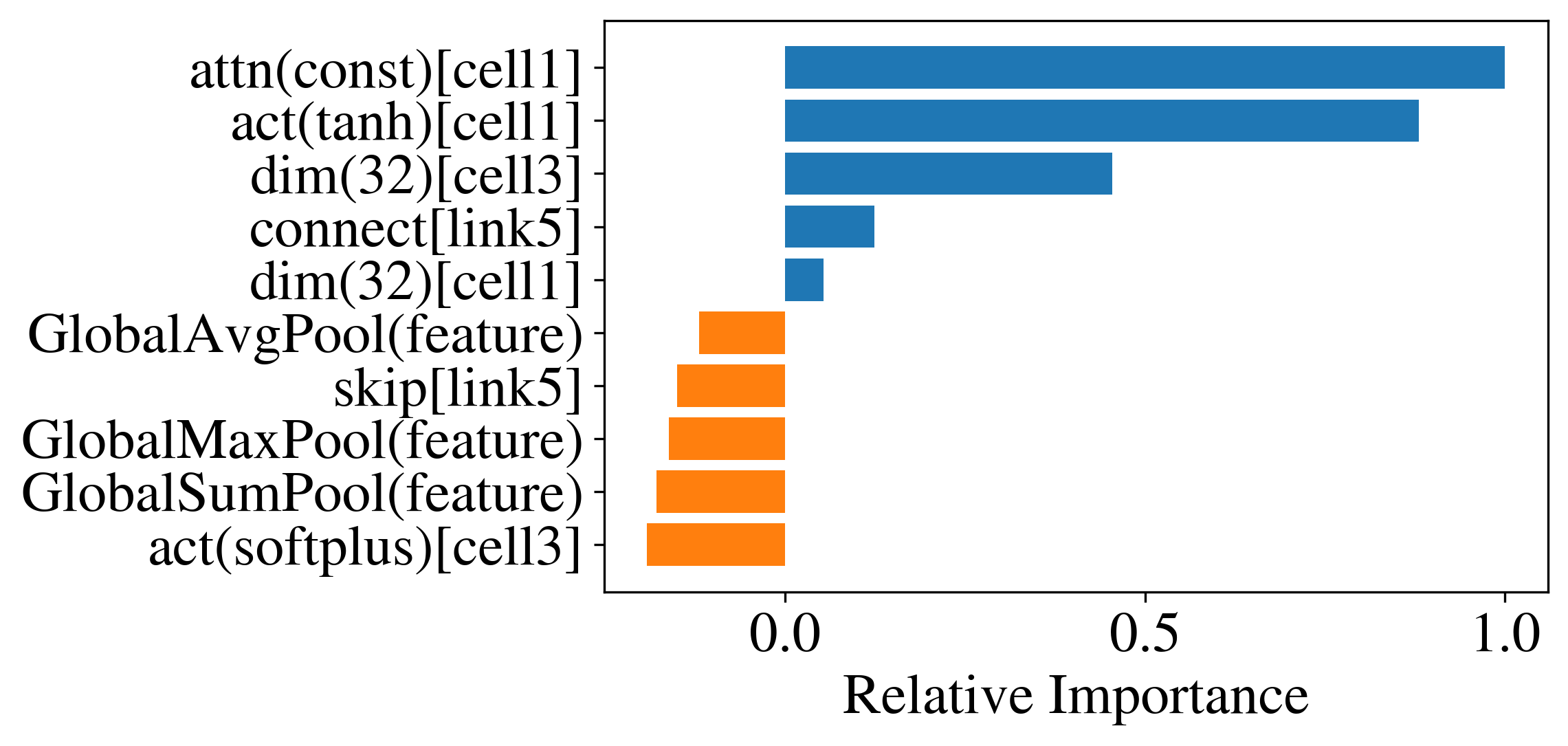}
  \caption{QM8}
  \label{fig:qm8_importance}
\end{subfigure}
\begin{subfigure}{0.6\columnwidth}
  \centering
  \includegraphics[width=\columnwidth]{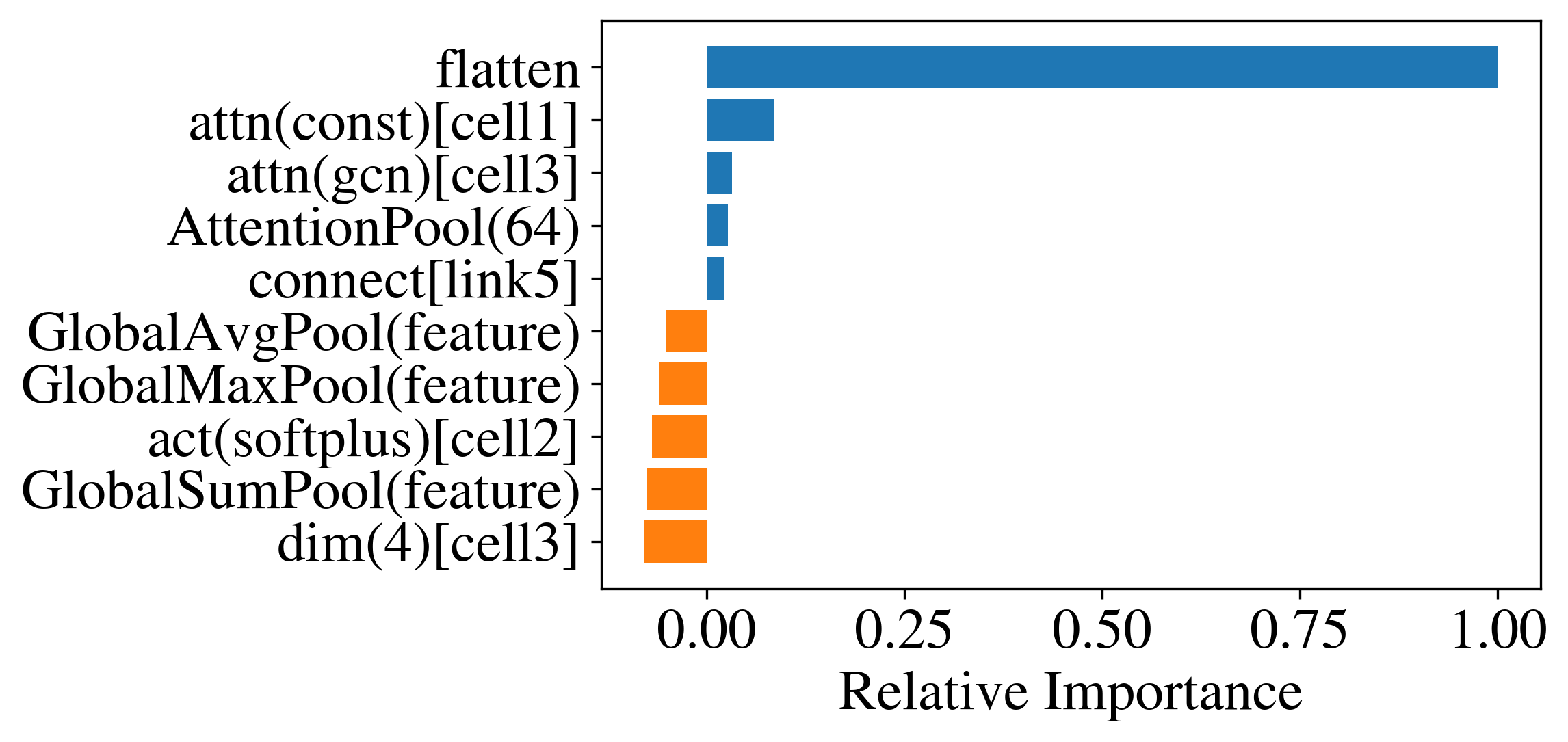}
  \caption{QM9}
  \label{fig:qm9_importance}
\end{subfigure}
\begin{subfigure}{0.6\columnwidth}
  \centering
  \includegraphics[width=\columnwidth]{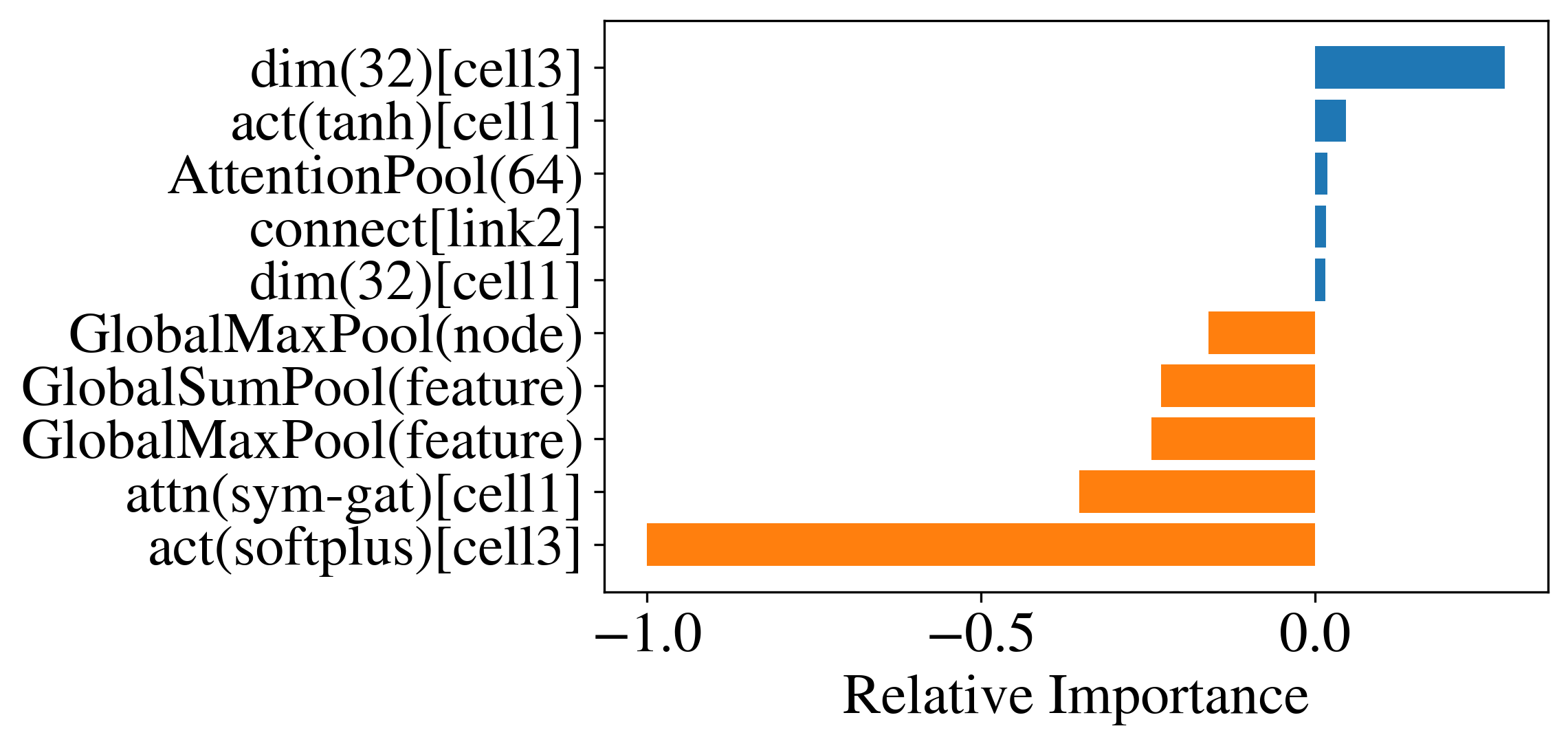}  
  \caption{ESOL}
  \label{fig:esol_importance}
\end{subfigure}
\begin{subfigure}{0.6\columnwidth}
  \centering
  \includegraphics[width=\columnwidth]{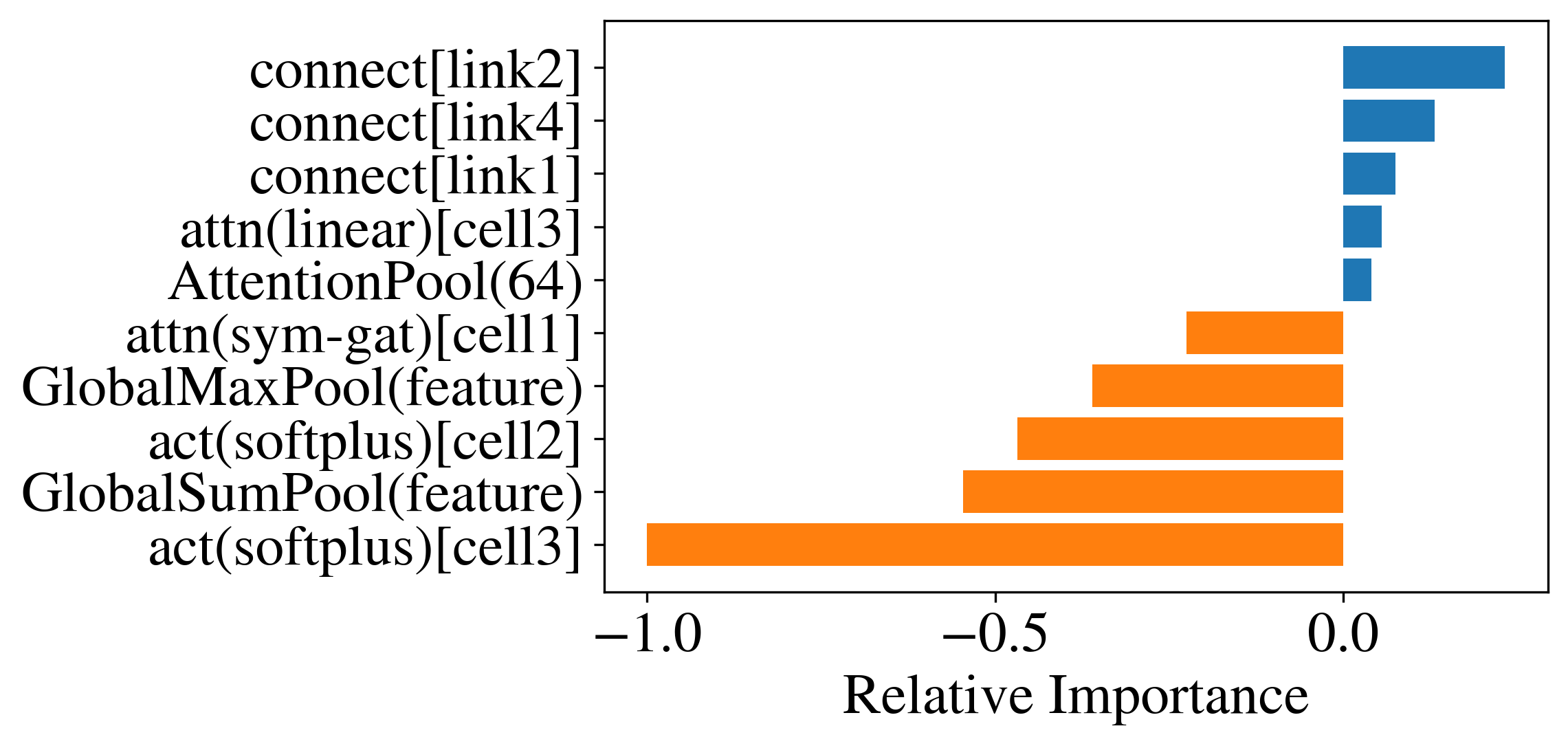}  
  \caption{FreeSolv}
  \label{fig:freesolv_importance}
\end{subfigure}
\begin{subfigure}{0.6\columnwidth}
  \centering
  \includegraphics[width=\columnwidth]{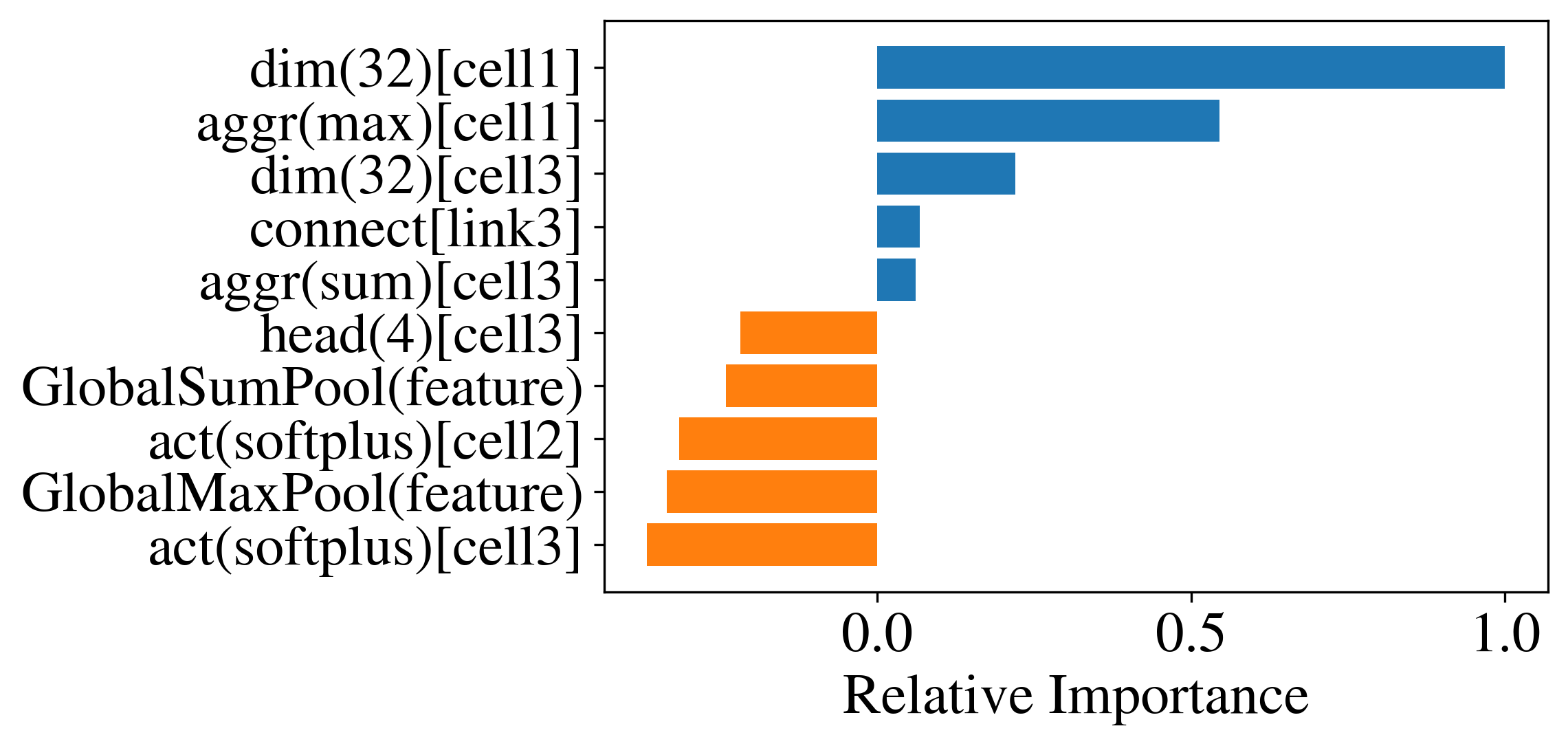}  
  \caption{Lipophilicity}
  \label{fig:lipo_importance}
\end{subfigure}
\caption{Importance plots on the test datasets. The operations with positive importance values increase the search reward in general, whereas negative importance values indicate that the corresponding operations hurt the architecture performance.}
\label{fig:importance}
\end{figure*}

We also have an operation vector $a$ and its corresponding contribution vector $b$. The overall relative importance vector $\mathcal{I}$ can be defined as $\mathcal{I} = \frac{1}{I}\sum_{i=1}^I a_i \odot b_i \label{eqn:tree-6}$,
where $I$ is the number of architecture samples, $a_i$ and $b_i$ are the operation and contribution vector of the $i$-th sample, respectively, and $\odot$ is the Hadamard product operation. 

The top five positive and negative impact operations for each dataset are shown in Fig.~\ref{fig:importance}. For QM7 and QM9, the flatten operation of the graph gather node is the most important. The constant attention method in the first MPNN node is significant for QM8. For ESOL and Lipophilicity, the state dimensions of 32 of the third and the first MPNN nodes are most critical, respectively. The skip-connection between the input and the second MPNN node is dominant for FreeSolv. In general, the skip-connection operations always increase the search reward, and Softplus activation functions tend to hurt the accuracy. Given the same search space, the different importance values indicate that each dataset requires a customized architecture. This finding reinforces the fact that manual design of customized architecture will be difficult and laborious but that NAS can greatly accelerate the process.


\subsection{Comparison between RE and RS}



Here, we compare RE to RS and study their scaling behavior. We show that RE outperforms RS with respect to search trajectory, number of architectures evaluated, and number of high-performing architectures. In the following study, we focused on running both search methods on 15, 30, and 60 nodes for QM8, QM9, ESOL, and Lipophilicity.


\subsubsection{Search trajectory}
From Figs.~\ref{fig:qm8_compare}, \ref{fig:qm9_compare}, \ref{fig:esol_compare} and \ref{fig:lipo_compare}, we observe that RE converges to a better search reward for all datasets with 15, 30, and 60 compute nodes. However, RS without any feedback mechanism fails to find high-performing architectures. Specifically, the RS search rewards for QM8, QM9, ESOL, and Lipophilicity are from -0.4 to -0.37, -0.51 to -0.47, -0.7 to -0.4, and -1.1 to -0.9, respectively; these rewards are worse than their corresponding RE results, which converge to between -0.35 and -0.31, -0.33 and -0.29, -0.31 and -0.3, and -0.82 and -0.64, respectively. In general, with an increasing number of nodes, RE search trajectories converge faster to a better search reward. For example, the RE search reward for QM8 reaches -0.36 at 53, 28, and 22 minutes with 15, 30, and 60 nodes, respectively. 

\subsubsection{Number of architectures evaluated}
The number of evaluations for RE and RS on a varying number of nodes for a fixed wall-clock time can be found in Table.~\ref{tab:scaling}. For all datasets, given the same number of nodes, the number of architectures evaluated is similar. For instance, RE evaluates 563 architectures with 30 nodes for QM8, whereas RS evaluates 540 architectures. With more nodes, however, the number of evaluations increases significantly. Specifically, RE for QM8 performs 219, 563, and 1490 evaluations with 15, 30, and 60 nodes. Similarly, RS for QM8 leads to 239, 540, and 1237 evaluations. The same trend is also seen for QM9, ESOL, and Lipophilicity. The difference in the number of evaluations between the two methods is caused by the training time variation for each architecture. For example, with 60 nodes, RE performs 253 more evaluations than RS does for QM8, but 81 less for QM9. The variation is because RE discovers more architectures that require less training time for QM8 than RS does. Unlike for QM8, RE for QM9 finds more architectures that are trained for a longer time than RS does.


\subsubsection{High-performing architectures discovered}
\begin{figure*}[h!]
\centering
\begin{subfigure}{0.85\columnwidth}
  \centering
  \includegraphics[width=0.9\columnwidth]{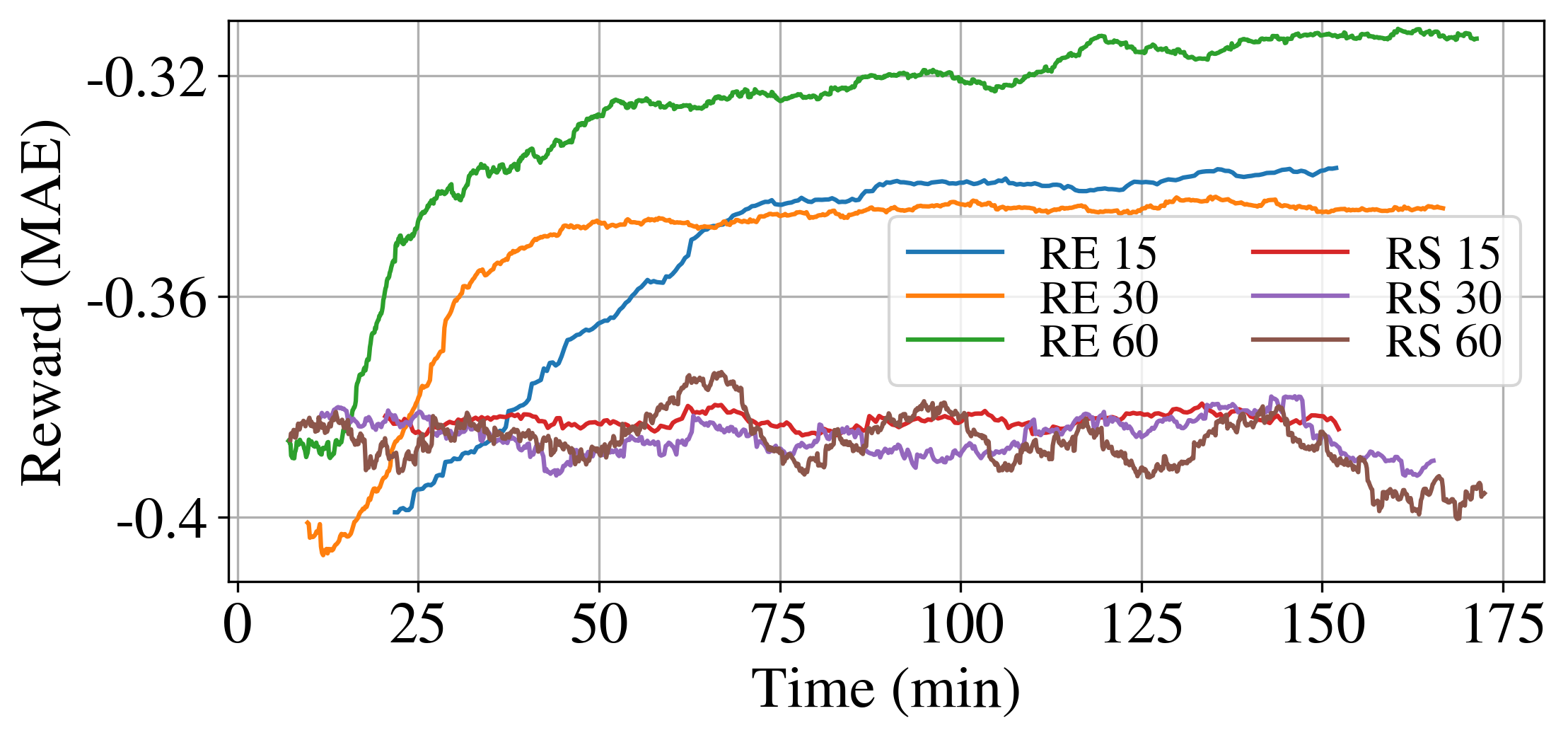}  
  \caption{QM8 search trajectory}
  \label{fig:qm8_compare}
\end{subfigure}
\begin{subfigure}{0.85\columnwidth}
  \centering
  \includegraphics[width=0.9\columnwidth]{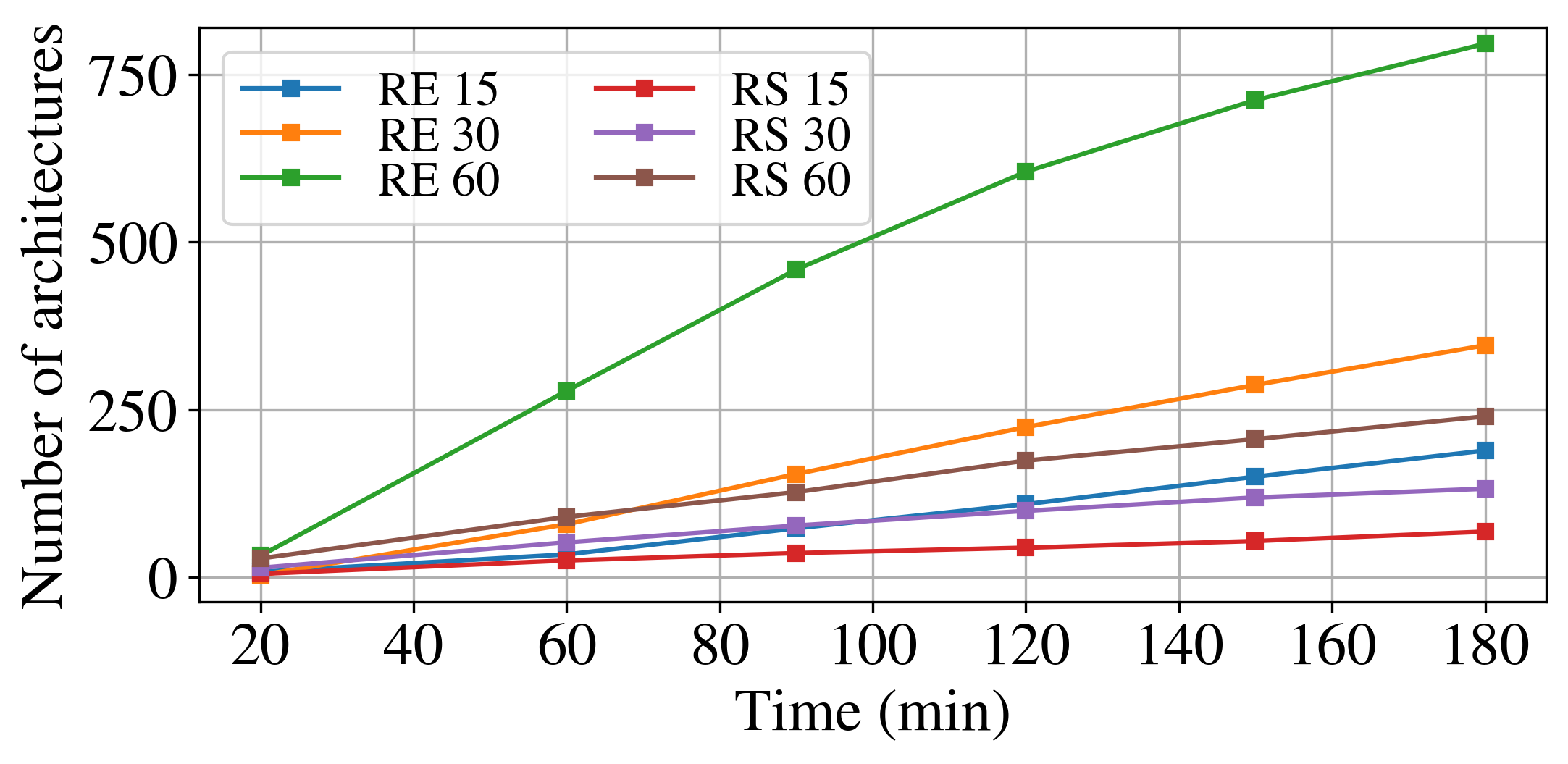}  
  \caption{QM8 high-performing architectures}
  \label{fig:qm8_unique}
\end{subfigure}

\begin{subfigure}{0.85\columnwidth}
  \centering
  \includegraphics[width=0.9\columnwidth]{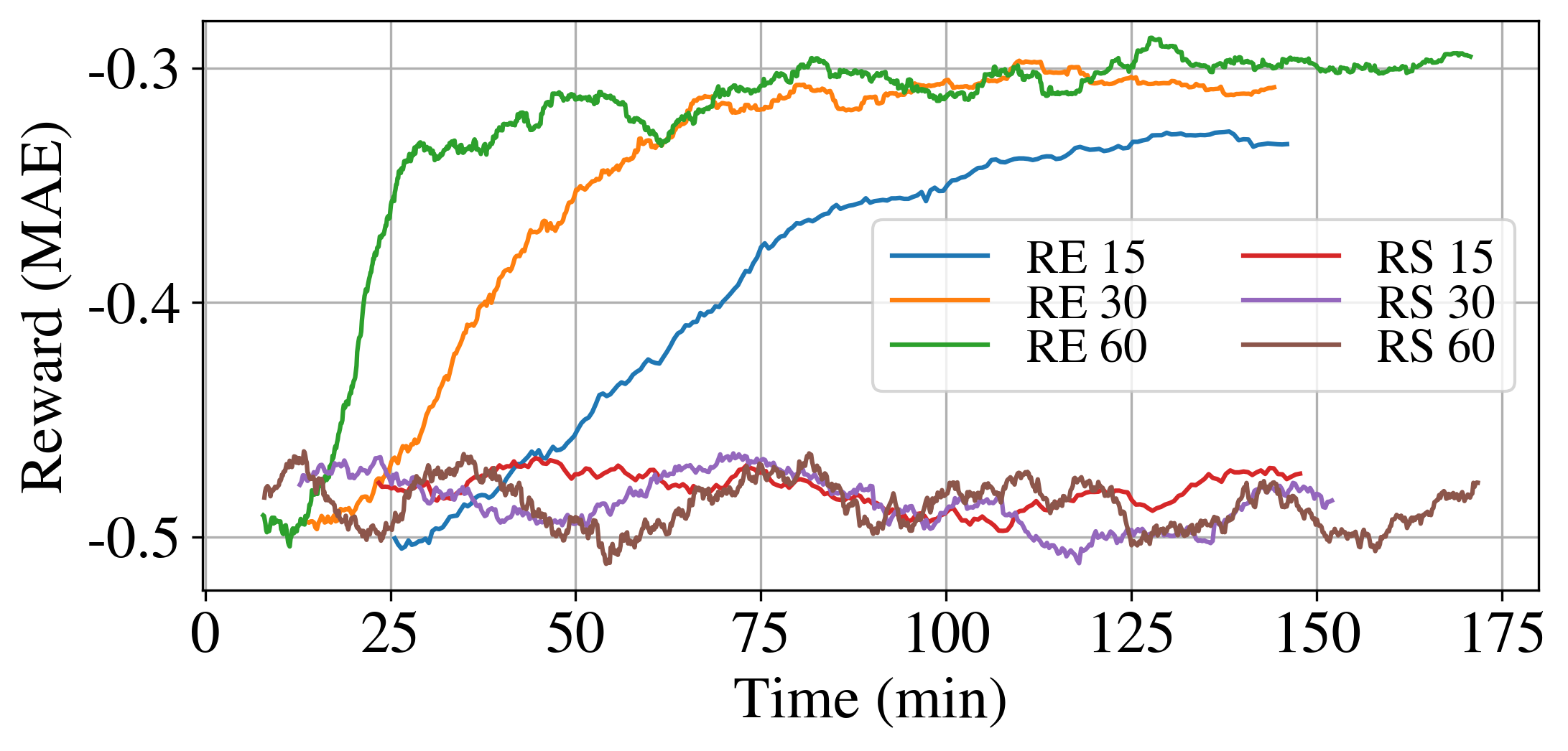}  
  \caption{QM9 search trajectory}
  \label{fig:qm9_compare}
\end{subfigure}
\begin{subfigure}{0.85\columnwidth}
  \centering
  \includegraphics[width=0.9\columnwidth]{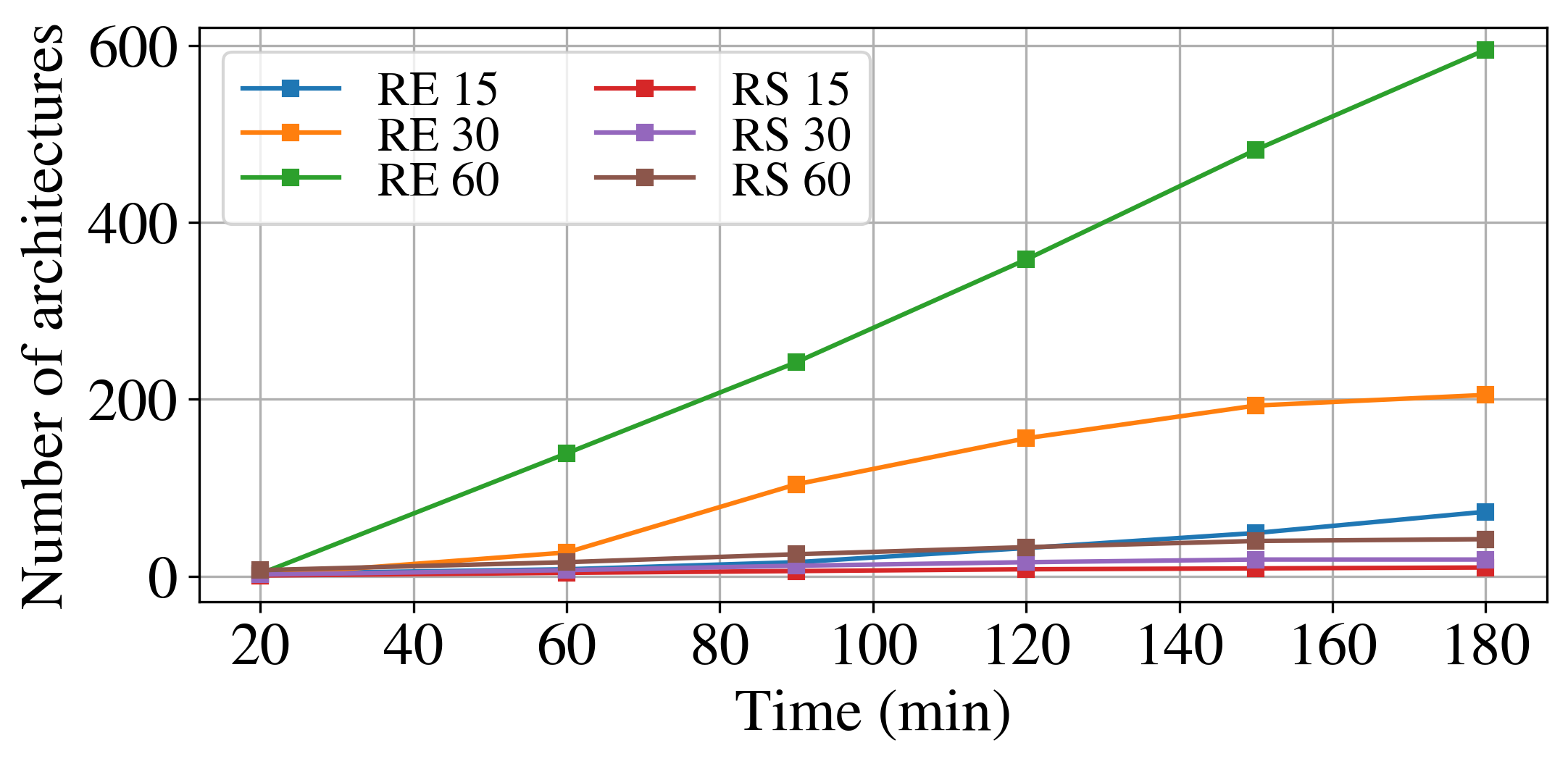}  
  \caption{QM9 high-performing architectures}
  \label{fig:qm9_unique}
\end{subfigure}

\begin{subfigure}{0.85\columnwidth}
  \centering
  \includegraphics[width=0.9\columnwidth]{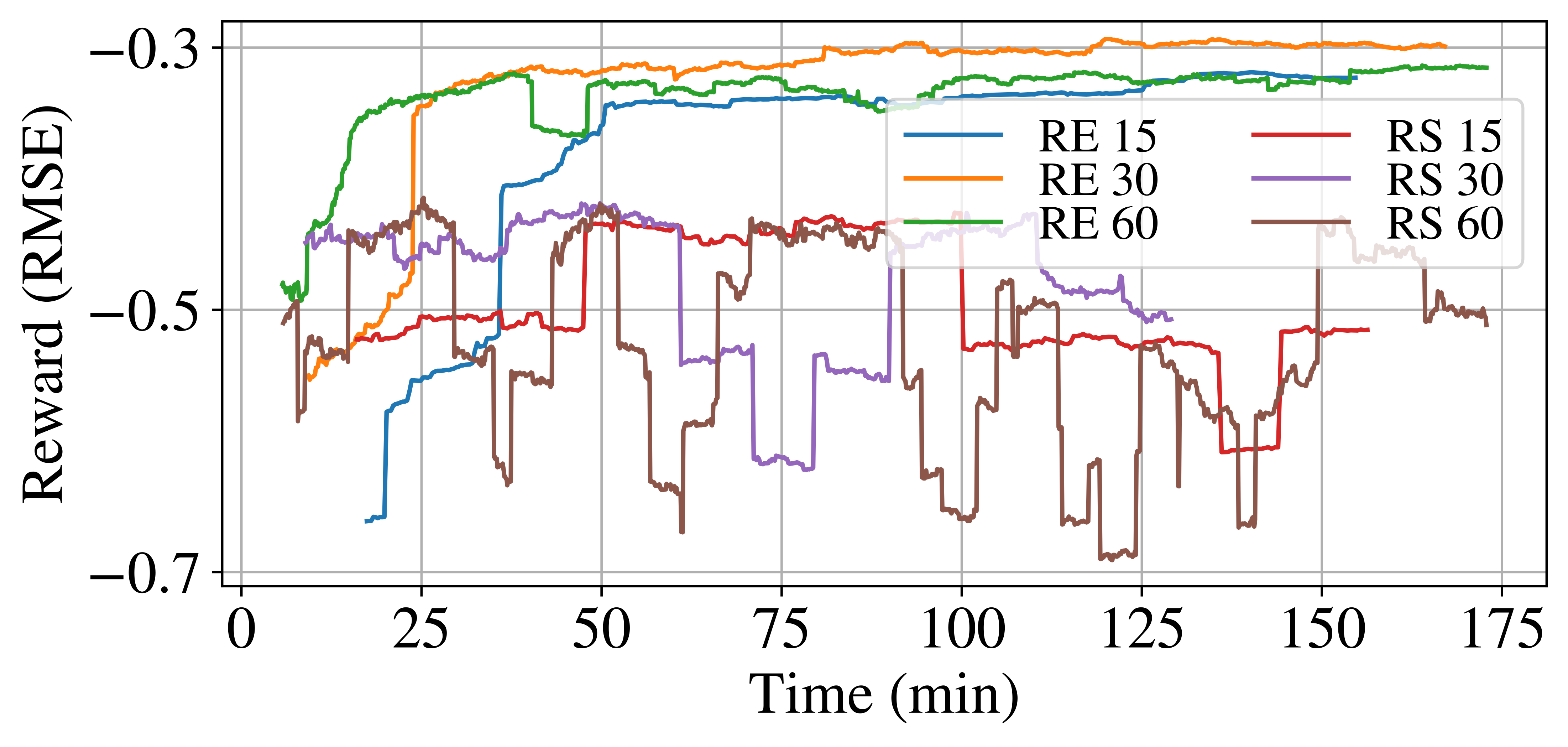}  
  \caption{ESOL search trajectory}
  \label{fig:esol_compare}
\end{subfigure}
\begin{subfigure}{0.85\columnwidth}
  \centering
  \includegraphics[width=0.9\columnwidth]{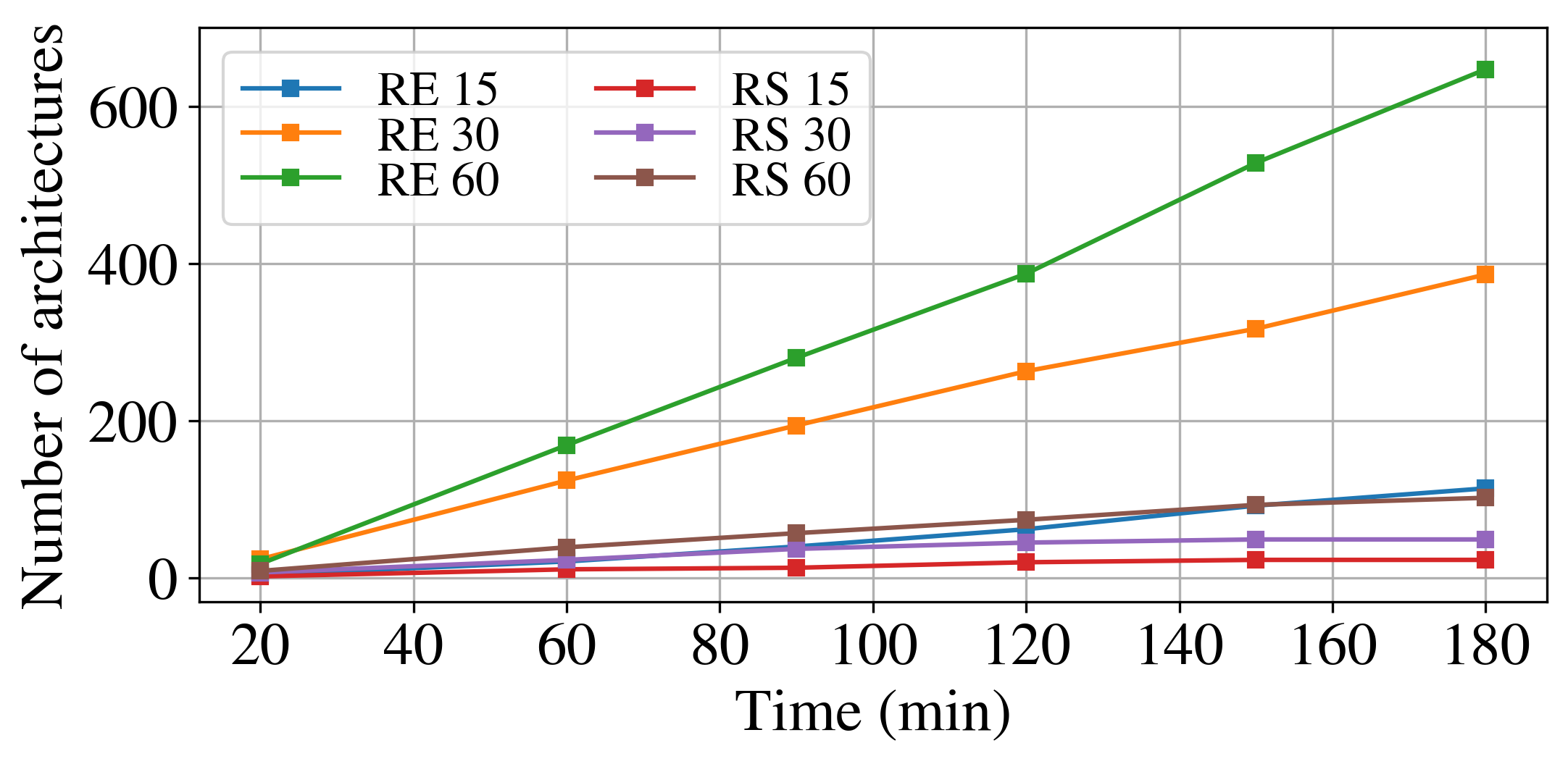}  
  \caption{ESOL high-performing architectures}
  \label{fig:esol_unique}
\end{subfigure}

\begin{subfigure}{0.85\columnwidth}
  \centering
  \includegraphics[width=0.9\columnwidth]{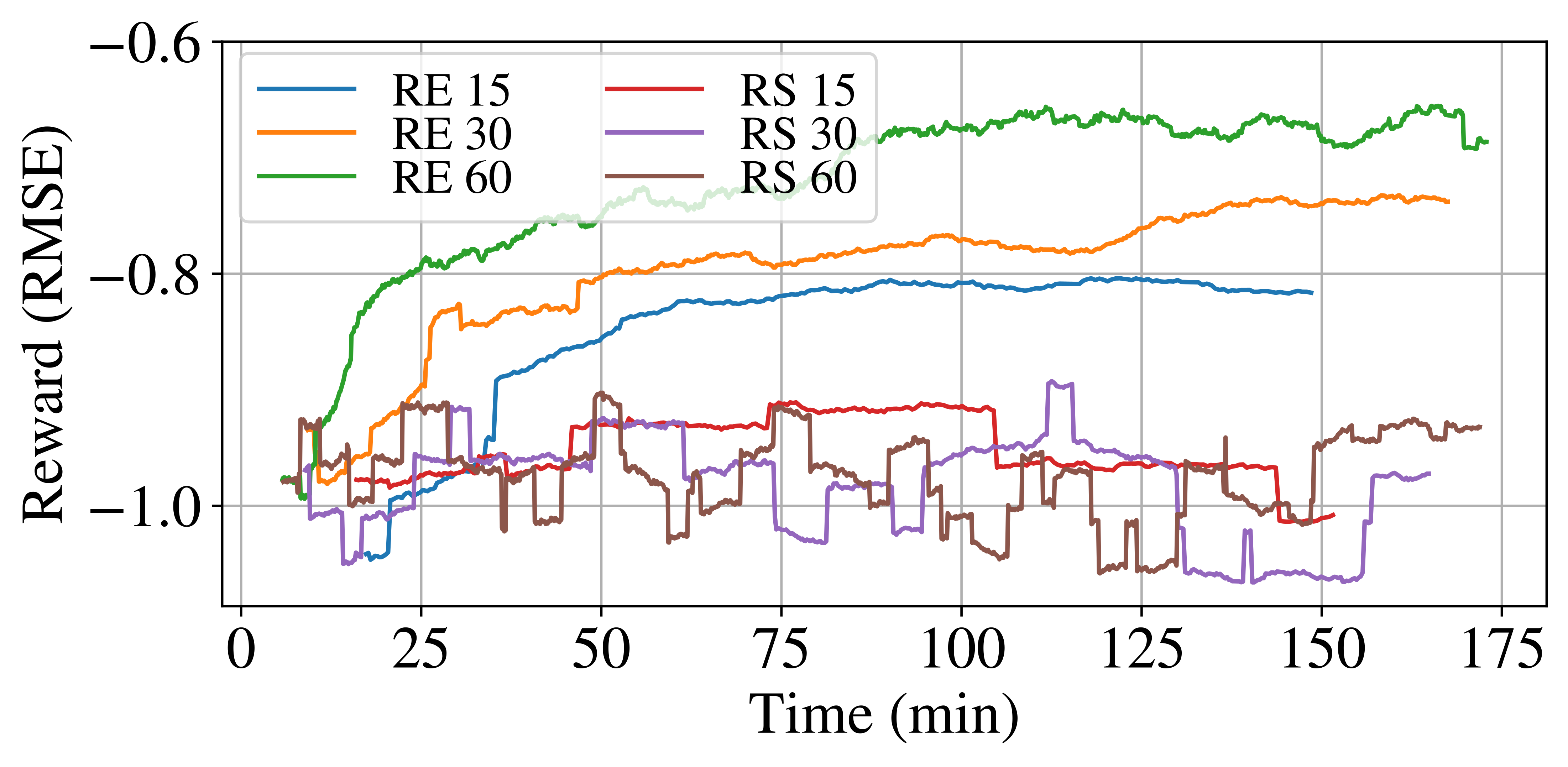}  
  \caption{Lipophilicity search trajectory}
  \label{fig:lipo_compare}
\end{subfigure}
\begin{subfigure}{0.85\columnwidth}
  \centering
  \includegraphics[width=0.9\columnwidth]{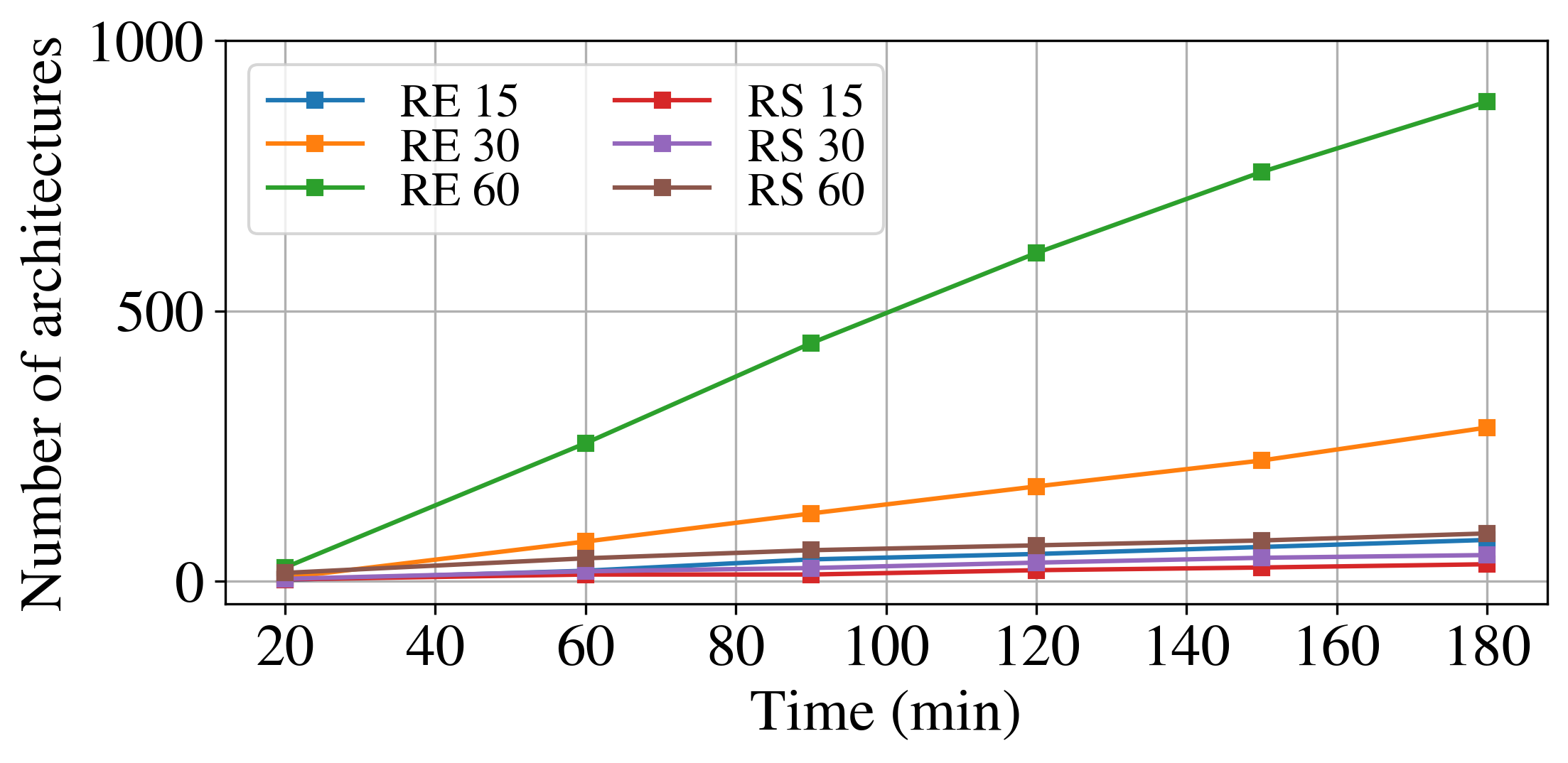}  
  \caption{Lipophilicity high-performing architectures}
  \label{fig:lipo_unique}
\end{subfigure}
\caption{Comparison of search trajectories between RE and RS and the temporal breakdown of the number of unique high-performing architectures discovered by RE and RS with 15, 30, and 60 compute nodes. On all the datasets, the RE search strategy is more effective at obtaining high-performing architectures.}
\label{fig:scaling}
\end{figure*}

In Figs.~\ref{fig:qm8_unique}, \ref{fig:qm9_unique}, \ref{fig:esol_unique}, and \ref{fig:lipo_unique}, we demonstrate the temporal breakdown of the number of unique high-performing architectures discovered by RE and RS on varying numbers of nodes. We define high-performing architectures as any architecture with a higher search reward than a given threshold value. The threshold values are -0.35, -0.3, -0.32, and -0.77 for QM8, QM9, ESOL, and Lipophilicity, respectively. The figure shows that the number of unique high-performing architectures grows considerably with a greater number of nodes for RE but not for RS. For example, the number of such architectures obtained by RE for QM8 at 180 minutes using 30 nodes is achieved by RE with 60 nodes in 70 minutes. We observe a similar trend for all datasets. Without a feedback mechanism, RS discovered a limited number of high-performing architectures, and the number increases insignificantly with more nodes. Specifically, after 180 minutes of search, RS for QM8 discovers 68, 132, and 240 high-performing architectures for 15, 30, and 60 nodes, respectively. In comparison, RE for QM8 finds 189, 346, and 796 architectures.


\begin{table}[tb]
\centering
\caption{Total number of evaluations for RE and RS on varying number of compute nodes}
\label{tab:scaling}
\begin{adjustbox}{width=\columnwidth}
\begin{tabular}{@{}c|cc|cc|cc|cc@{}}
\toprule
 & \multicolumn{2}{c|}{QM8} & \multicolumn{2}{c|}{QM9} & \multicolumn{2}{c|}{ESOL} & \multicolumn{2}{c}{Lipo} \\ \midrule
No. of nodes & RE & RS & RE & RS & RE & RS & RE & RS \\
15 & 219 & 239 & 282 & 295 & 284 & 337 & 356 & 368 \\
30 & 563 & 540 & 726 & 694 & 742 & 775 & 773 & 667 \\
60 & 1490 & 1237 & 1393 & 1474 & 1880 & 1647 & 1851 & 1748 \\ \bottomrule
\end{tabular}
\end{adjustbox}
\end{table}

\section{Related Work}

For a detailed review of NAS for graphs, we refer the reader to \cite{he2019automl}. Here, we review prior works related to GNNs and NAS.

\textbf{Graph Neural Networks.} A wide variety of GNNs have been invented to study node embeddings. Most of these can be viewed as MPNNs with information passed from one node to another. GCN \cite{kipf2016semi} focuses on node information only, and the information passed between any two nodes is equally weighted. GAT \cite{velivckovic2017graph} calculates an attention coefficient as a weight for information passed between two nodes based on the properties of neighboring nodes. By incorporating edge features into message passing, RGCN \cite{schlichtkrull2018modeling}, GGNN \cite{li2015gated}, and LanczosNet \cite{liao2019lanczosnet} have better performance but lack the flexibility for continuous edge information. The formal MPNNs \cite{gilmer2017neural} train the edge features to govern the node information passing. Our work combines edge information and attention mechanisms to further guide the node information passing.

\textbf{Neural Architecture Search.} Several studies of automated GNN architecture searches have been conducted for a diverse set of applications. GraphNAS, proposed in \cite{gao2019graphnas}, applies RNN directly to search GNN structures; the whole neural architecture is sampled and reconstructed in GraphNAS. AGNN \cite{zhou2019auto}, in contrast, explores the offspring architecture by modifying only a specific operation class; the best architectures are retained to provide a good start for architecture modification. Both  works contain state dimension, attention function, attention head, aggregate functions, and activation functions in the search space. The merge or combine function is similar to the update function in MPNNs. However, both works lack support for edge features. Our work not only embeds edge information but also enables skip-connection and modification to the graph gather node.

\section{Conclusion and Future Work}
We introduced a neural architecture for the automated development of stacked MPNNs to predict the molecular properties of small molecules from the MoleculeNet benchmark dataset. This is the first work that employs NAS for MPNNs and establishes the first graph-related search space to incorporate edge features with skip connections. 


We developed a MPNN search space that comprises MPNN nodes and skip connections. The MPNN node comprises various state dimensions, attention functions, attention head, aggregate function, activation functions, update functions, and gather operations. As a search strategy, we utilized RE, an asynchronous evolutionary algorithm within DeepHyper, an open-source automated machine learning package. We compared RE with RS in DeepHyper and showed that RE outperforms RS in terms of the number of high-performing architectures discovered. In addition, RE achieves architectures with better performance in a shorter wall-clock time and matches the scalability of the completely asynchronous RS. We compared the manually designed MoleculeNet benchmarks with the best architectures obtained using RE that was retrained for a longer number of epochs. We showed that the automatically designed architecture outperformed the baselines with respect to loss on the validation and test data. 

Using a random-forest-based nonparamteric sensitivity analysis method, we analyzed the importance of the different choices in the stacked MPNN search space. The results from the NAS showed the importance of MPNN customization. Given the same search space, search method, and evaluation strategy, we showed that the choices differs significantly from dataset to dataset.  


Our future work will seek to overcome the limitations of processing large molecules by transforming the search space from Keras to TensorFlow 2.0, allowing us to eliminate the padding of input data. Moreover, we will develop a search space for graph autoencoders for generative modeling. The results from this work also show promise  for continued investigation into AutoML-enhanced, data-driven surrogate discovery for scientific machine learning.

\section*{Acknowledgment}
This material is based on work supported by the U.S.\ Department of Energy 
(DOE), Office of Science, Office of Advanced Scientific Computing Research, under
Contract DE-AC02-06CH11357. We gratefully acknowledge the computing resources provided on Bebop, a high-performance computing cluster operated by the Laboratory Computing Resource Center at Argonne National Laboratory.

\bibliographystyle{IEEEtran}
\bibliography{IEEEabrv, GNN_NAS.bib}

\end{document}